\pdfoutput=1
\documentclass[11pt]{article}

\usepackage[]{acl}

\usepackage{times}
\usepackage{latexsym}
\usepackage{cite}
\usepackage{amsmath,amssymb,amsfonts}
\usepackage{subcaption}
\usepackage{algorithmic}
\usepackage{graphicx}
\usepackage{textcomp}
\usepackage{xcolor}
\usepackage{textcomp}
\usepackage{mathtools}  
\usepackage{dsfont}
\usepackage{hyperref}
\usepackage{booktabs}
\usepackage{multirow}

\DeclareMathOperator*{\E}{\mathbb{E}}

\newcommand{\secref}[1]{Section~\ref{#1}}
\newcommand{\figref}[1]{Figure~\ref{#1}}
\newcommand{\subfig}[1]{\textit{#1}}
\newcommand{\tabref}[1]{Table~\ref{#1}}

\newcommand{\appendixref}[1]{Appendix~\ref{#1}}

\newcommand{\ie}{\textrm{i.e.,}}
\newcommand{\eg}{\textrm{e.g.,}}

\def\blog{Blog}
\def\blogsmall{Blog10}
\def\blogbig{Blog50}
\def\imdb{IMDb62}
\def\enron{Enron100}
\def\turingbench{TuringBench}

\def\bertbase{BERT}
\def\bertcl{Contra-BERT}
\def\debertabase{DeBERTa}
\def\debertacl{Contra-DeBERTa}
\def\turingtest{\emph{Turing Test}}
\def\method{Contra-X}

\def\Ai{$A_i$}
\def\Aj{$A_j$}

\usepackage[T1]{fontenc}
\usepackage[utf8]{inputenc}
\usepackage{enumitem}
\usepackage{microtype}

\title{Whodunit? Learning to Contrast for Authorship Attribution}

\author{Bo Ai$^{1}$, Yuchen Wang$^{1}$, Yugin Tan$^{1}$, Samson Tan$^{2}$\thanks{{ } Work done at the National University of Singapore.} \\
  $^{1}$School of Computing, National University of Singapore \\
  $^{2}$AWS AI Research \& Education 
}

\newcommand\blfootnote[1]{% 
\begingroup 
\renewcommand\thefootnote{}\footnote{#1}% 
\addtocounter{footnote}{-1}% 
\endgroup 
}

\begin{document}
\maketitle
% \begin{abstract}
% Authorship attribution is the task of identifying the author of a given text. Most existing approaches use manually designed features that capture a dataset's content and style, but these dataset-dependent approaches yield inconsistent performance across different corpora. In this work, we propose using generic pre-trained language representations as a basis and fine-tuning these representations with contrastive learning to learn author-specific features (Contra-X). We show that Contra-X advances the state-of-the-art on multiple human and machine authorship attribution benchmarks, enabling improvements of up to 6.8\%. We also show that Contra-X consistently outperforms cross-entropy fine-tuning across different data regimes. Crucially, we present qualitative and quantitative analyses of these improvements. Our learned representations form highly separable clusters for different authors. However, we find that Contra-X improves overall accuracy at the cost of sacrificing performance for some authors. Resolving this tension will be an important direction for future work. To the best of our knowledge, we are the first to analyze the effect of integrating contrastive learning with cross-entropy fine-tuning for authorship attribution.
% \blfootnote{Code and datasets are available at \textcolor{magenta}{\url{https://github.com/BoAi01/Contra-X.git}}}
% \end{abstract}

\begin{abstract}
Authorship attribution is the task of identifying the author of a given text. The key is finding representations that can differentiate between authors. Existing approaches typically use manually designed features that capture a dataset's content and style, but these approaches are dataset-dependent and yield inconsistent performance across corpora. In this work, we propose \textit{learning} author-specific representations by fine-tuning pre-trained generic language representations with a contrastive objective (Contra-X). We show that Contra-X learns representations that form highly separable clusters for different authors. It advances the state-of-the-art on multiple human and machine authorship attribution benchmarks, enabling improvements of up to 6.8\% over cross-entropy fine-tuning. However, we find that Contra-X improves overall accuracy at the cost of sacrificing performance for some authors. Resolving this tension will be an important direction for future work. To the best of our knowledge, we are the first to integrate contrastive learning with pre-trained language model fine-tuning for authorship attribution.
\blfootnote{Implementation and datasets are available at \textcolor{magenta}{\url{https://github.com/BoAi01/Contra-X.git}}}
\end{abstract}

\section{Introduction}
% Structure
% \begin{itemize}
%     \item Hook: What is the state of the world/field/community today, what is the significance of this problem and how does it relate to the world? Can tie in to machine generated text detection, how it is a special case of AA.
%     \item How will the world be changed if we make progress on this problem? (Why should the reader continue reading)?
%     \item What are the gaps in existing methods? X, Y, Z have done a, b, c, BUT what about d? What did they miss? What are the implications?
%     \item Some lead-in/background to your method. However, its impact on authorship attribution has not been investigated. 
%     \item Hence, we contribute/investigate the effect of ..., empirically demonstrate its efficacy on multiple datasets, human and machine-generated texts, while being more sample efficient with the same number of training steps.
%     \item also propose metrics to analyze how constrastive learning improves authorship attribution performance.
% \end{itemize}

\begin{figure}[t]
    \centering
    \setlength{\tabcolsep}{2pt} % Default value: 6pt
    \renewcommand{\arraystretch}{2} % Default value: 1
    \begin{tabular}{ c c }
    \includegraphics[width=0.45\columnwidth]{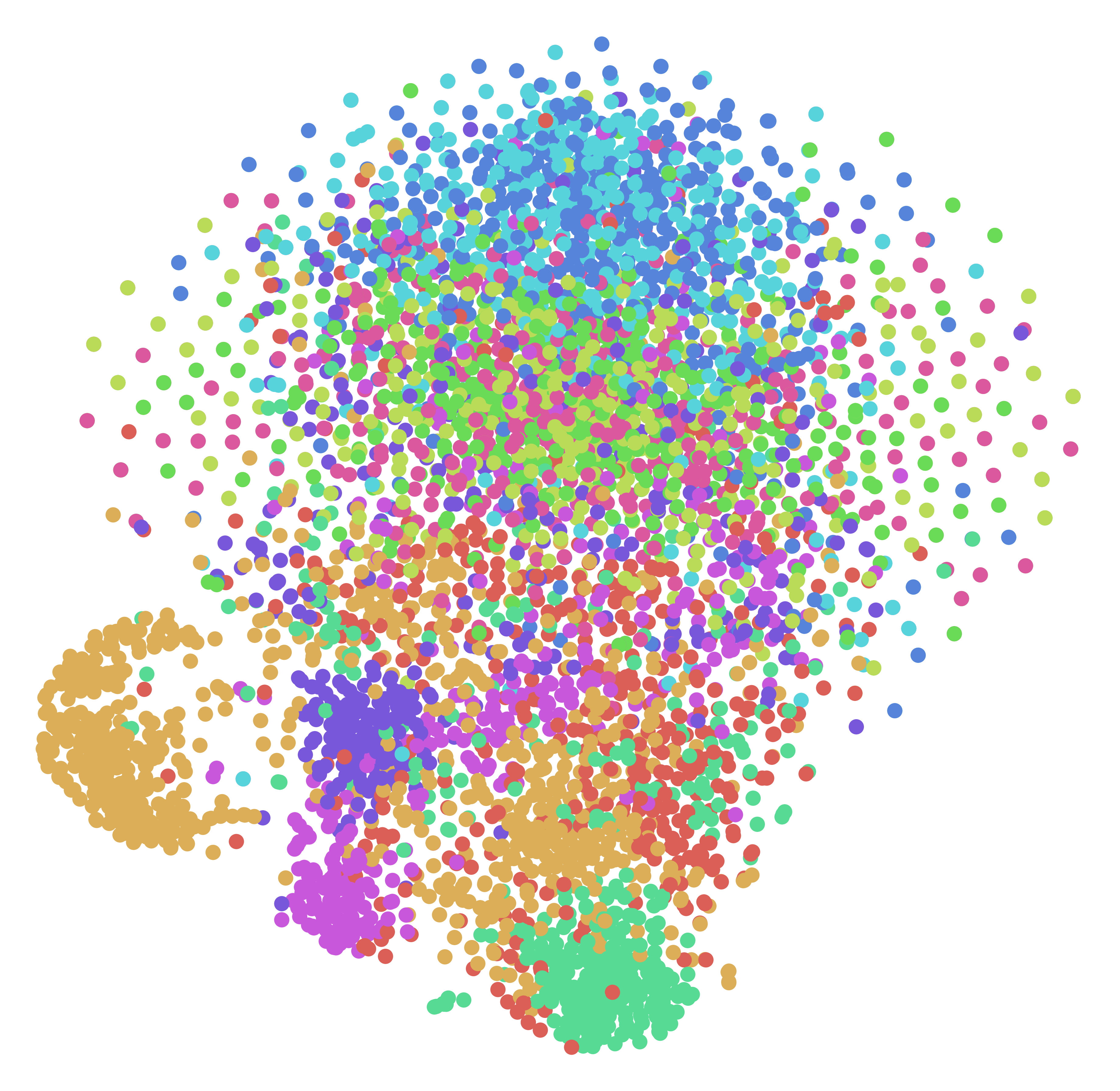} \label{fig:tsne_base} &
    \includegraphics[width=0.45\columnwidth]{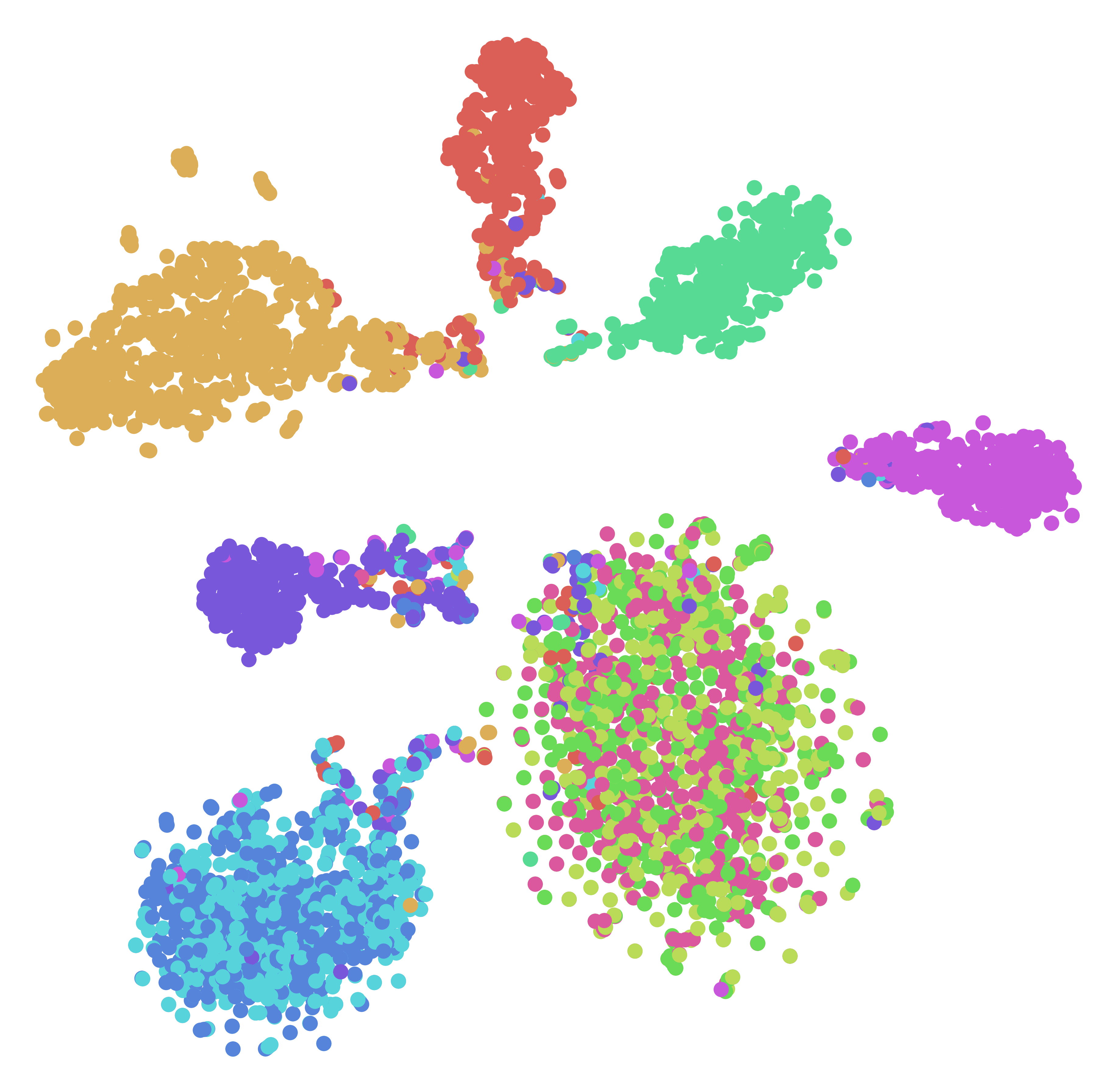} \label{fig:tsne_ours} \\
    \subfig(a) \bertbase{} & \subfig(b) \bertcl{} \\ 
    % \includegraphics[width=0.48\textwidth]{figs/matrix_baseline_50_cropped.png} \label{fig:conf_base} &
    % \includegraphics[width=0.48\textwidth]{figs/matrix_proposed_50_cropped.png}
    % \label{fig:conf_ours} \\
    % \subfig(c) \bertbase{} confusion matrix & \subfig(d) \bertcl{} confusion matrix \\ 
    \end{tabular}
    \caption{t-SNE visualization of the fine-tuned representations (a: baseline; b: Contra-X). Each color denotes one author in the \blogsmall{} dataset. Our contrastive method effectively creates a tighter representation spread for each author and increased separation between authors. Best viewed in color. }
    \label{fig:tsne}
\end{figure}

Authorship attribution (AA) is the task of identifying the author of a given text. AA systems are commonly used to identify the authors of anonymous email threats \citep{forensic_1} and  historical texts \citep{mendenhall1887characteristic}, and to detect plagiarism \citep{forensics_plagiarism}. With the rise of neural text generators that are able to create highly believable fake news \citep{grover}, AA systems are also increasingly employed in machine-generated-text detection \citep{detection_survey}. When performed on texts generated by human and machine writers, AA can also act as a type of \turingtest{} for Natural Language Generation \citep{turing_benchmark, aa_nlg_2}. 

% Authorship attribution (AA) is the task of identifying the author of a given text. AA systems can be used for plagiarism detection \citep{forensics_plagiarism}, email forensics \citep{forensic_1}, and even the detection of machine-generated text \citep{detection_survey}. The third is particularly pertinent as neural text generators are increasingly used to create highly believable fake news \citep{grover}. %IFor example, artificial text detection \citep{detection_survey} can be viewed as a binary authorship attribution in which the only two authors are a human and a machine. Moreover, the task can also serve as the \turingtest{} for Neural Language Generation (NLG) \citep{turing_benchmark, aa_nlg}. Do NLG models write differently and how do they differ from human writing? We can perform authorship attribution on their writing to assess the similarity. Hence, advances on the authorship attribution task not only benefits the practical applications, but also contributes to combating fake machine-generated contents and promots the advances in NLG. 

Traditional AA methods design features that characterize texts based on their content or writing style \citep{style_aware, syntax_cnn, style_ngram, topic_or_style}. However, the features useful for distinguishing authors are often dataset-specific, yielding inconsistent performance under varying conditions \citep{topic_or_style}. In contrast, learning features from large corpora of data aims to produce general pre-trained models \citep{bert} that improve performance on many core natural language processing (NLP) tasks, including AA \citep{bertaa}. However, it is unclear if basic fine-tuning makes full use of the information in the training data. We seek to augment the learning process.

Contrastive learning is a technique that pulls similar samples close and pushes dissimilar samples apart in the representation space \citep{simcse}. It has proven useful in tasks that require distinguishing subtle differences \citep{cl_multiview, cl_speech}. This makes it highly suited to encouraging the learning of distinct author subspaces. However, no prior work has investigated its relevance to the AA task. To this end, we seek to understand its impact on the learning of author-specific features under the supervised learning paradigm.

%In this work, we seek to investigate contrastive learning as a method to learn discriminative features for the author authorship attribution task. 

To achieve this, we integrate \textbf{\uppercase{contra}}stive learning with \textbf{\uppercase{cross}}-entropy fine-tuning (\textbf{\method{}})  and demonstrate its efficacy via evaluation on multiple AA datasets. Unlike previous AA work, we evaluate the proposed approach not only on human writing corpora but also on machine-generated texts. There are three major reasons. First, this can show that our approach is generic to writer identity and dataset composition. Second, performing AA on human and machine authors reflects the increased importance of identifying machine-generated text sources. Third, this potentially reveals information about how differently machines write compared to humans. In addition, we study the performance of our method under different data regimes. We find \method{} to consistently improve model performance and yield distinct author subspaces. Finally, we analyze the performance gains vis-\`a-vis a number of AA-specific stylometric features. To the best of our knowledge, we are the first to incorporate contrastive learning into large language model fine-tuning for authorship attribution.

\section{Related Work} \label{sec:related_work}

\paragraph{Authorship attribution.} AA techniques fall under two broad categories: feature-based and learning-based approaches. The former involves hand-crafting features relevant for identifying authors \citep{topic_or_style}; the latter exploits large-scale pretraining to learn text representations. 

We note that feature-based approaches are investigated in two streams of work. One stream benchmarks on public datasets such as \imdb{} \citep{imdb62} and \blog{} \citep{blog}. The various features proposed include term frequency-inverse document frequency (TF-IDF) \citep{rahgouy2019cross}, letter and digit frequency \citep{topic_or_style}, and character n-grams \citep{char_ngram}. The other stream is the PAN shared task of authorship identification. These methods typically use multiple features such as n-grams \citep{pan1,pan2,pan3,pan5,pan6} in an ensemble. The two streams share similar technical ideas and developments.

However, feature-based approaches require dataset-specific engineering \citep{topic_or_style} and their performance does not scale with more data In contrast, learning-based approaches learn representations completely from data. BertAA \citep{bertaa} shows that a simple fine-tuning of pre-trained language models can outperform classical approaches by a clear margin. However, purely cross-entropy fine-tuning may not directly address the challenge of learning distinctive representations for different authors. Thus, we propose to incorporate contrastive learning, which explicitly enforces distance constraints in the representation space.

\paragraph{Contrastive learning.} Contrastive learning aims to learn discriminative features by pulling semantically similar samples close and pushing dissimilar samples apart. This encourages the learning of highly separable features that can be easily operated on by a downstream classifier. Unsupervised contrastive learning has been used to improve the robustness and transferability of speech recognition \citep{cl_speech} and to learn semantically meaningful sentence embeddings \citep{simcse}. It has also been combined with supervised learning for intent detection \citep{contrastive_pt_ft}, punctuation restoration \citep{supervised_cl_punctuation}, machine translation \citep{supervised_cl}, and dialogue summarization \citep{cl_dialogue}. However, to the best of our knowledge, we are the first to study its efficacy and limitations on authorship attribution.

\paragraph{Detection of machine-generated text.} Modern natural language generation (NLG) models can generate texts indistinguishable from human writings \citep{gpt3, grover}. With the potential for malicious use such as creating fake news \citep{ntg_impact}, detecting machine-generated text is increasingly important. This binary classification task can be extended to a multi-class AA task including both humans and NLG authors. This task can therefore identify not just machine text but also its particular source. Further, \citet{turing_benchmark} proposes that this serves as a \emph{Turing Test} to assess the quality of NLG models. Hence, we evaluate our approach on both human corpora and the human-machine dataset \turingbench{}, and show that our approach is generic to author identity and dataset composition. 

% Also known as artificial or neural-generated text \textit{detection} \citep{detection_survey, detection_topology}, this is a special type of authorship attribution. 

% Existing detectors perform inconsistently on texts from different generators \citep{turing_benchmark}. 
% We investigate how contrastive learning can produce a stable and performant detector as well as its potential limitations. 

% unlike typical previous work on authorship attribution task that only evaluates the method on human texts, we also present results on 

\section{Methodology} \label{sec:method}

\subsection{Problem formulation}
Authorship attribution is a classification task where the input is some text, $t$, and the target is the author, $a$. Formally, given a corpus $\mathcal{D}$, where each sample is a text-author pair $\langle t, a \rangle$, we aim to learn a predictor, $p$, that maximizes the prediction accuracy:
\begin{equation}
    Acc = \E_{\langle{}t, a\rangle{} \in D} \mathds{1}_{argmax(p(t))=a}
\end{equation}

Conventionally, this is achieved by optimizing a surrogate cross-entropy loss function via mini-batch gradient descent. Assuming we have a mini-batch containing $N$ texts $\{t_i\}_{i=1:N}$ and corresponding authors $\{a_i\}_{i=1:N}$, the loss function is: 
\begin{equation} \label{ce_loss}
    \mathcal{L}_{CE} = - \sum_{i} a_i \log(p(t)_{a_i})
\end{equation}

However, we hypothesize that $\mathcal{L}_{CE}$ does not adequately reflect the key challenge of the task, which is to learn highly discriminative representations for the input texts such that authorship can be clearly identified. Thus, we propose to augment the loss with a contrastive learning objective.

\subsection{\method{} for Authorship Attribution} \label{sec:cl}
We conjecture that the key to the authorship attribution task is to learn highly author-specific representations that capture each author's characteristics. Specifically, this requires representations to be similar for samples from the same authors, but distinct for samples from different authors. We adopt two specific strategies to achieve this goal: 
\begin{itemize}[leftmargin=\parindent]\itemsep0em
    \item Unlike most previous work that hand-crafts features and then learns a predictor $p$ from scratch, we fine-tune the general representations acquired from the large-scale unsupervised pre-training. Specifically, we decompose $p$ as $p = \phi \circ h$ where $\phi$ is the pre-trained language model and $h$ is a classifier layer. As shown by BertAA \citep{bertaa}, the learned representation is a decent starting point for the task.
    \item However, different from BertAA that fine-tunes the model $p = \phi \circ h$ with cross-entropy, we use an additional contrastive objective to encourage $\phi$ to capture the idiosyncrasies of each author. We conjecture that this can better exploit the information in the training data. 
\end{itemize}
% This is to avoid the dataset-specific feature engineering, and such learned representations have stronger expressive 

Intuitively, the contrastive loss encourages the model to \textbf{\emph{maximize}} the representational similarity of texts written by the same author, \ie{} positive pairs, and \textbf{\emph{minimize}} the representational similarity of texts written by different authors, \ie{} negative pairs. Formally, given a mini-batch containing $N$ texts $\{t_i\}_{i=1:N}$ and their authors $\{a_i\}_{i=1:N}$, we feed them into a pre-trained language model $\phi$ to obtain a batch of embeddings $\{e_i\}_{i=1:N}$, where $e_i = \phi(t_i)$. Embeddings of two samples by the same author $\langle e_{i}, e_{j} \rangle_{a_i = a_j}$ are a positive pair, and embeddings of two samples by different authors $\langle  e_{i}, e_{j} \rangle_{a_i \neq a_j}$ are a negative pair. We construct a similarity matrix $\mathcal{S}$ in which the entry $(i, j)$ denotes the pairwise similarity between $e_{i}$ and $e_{j}$. Formally, 
\begin{equation}
    \mathcal{S}_{i, j}  = \cos(e_i, e_j) = \frac{e_i \cdot e_j}{\|e_i\| \|e_j\|} 
\end{equation}

To encourage the abovementioned pairwise constraints, we define the contrastive objective as: 

\begin{align} \label{eqn:cl_term}
    \mathcal{L}_{CL} = & -\sum_{i} \log(\frac{\sum_{a_i = a_j} \exp(\cos(e_i, e_j)/\tau)}{\sum_{k}  \exp(\cos(e_i, e_k)/\tau)}) \nonumber \\
    = & -\sum_i \log(\frac{\sum_{a_i = a_j} \exp(\mathcal{S}_{i,j}/\tau)}{\sum_k \exp(\mathcal{S}_{i,k}/\tau))}),
\end{align}
where $\tau$ is the temperature. The loss could be viewed as applied on a softmax distribution to maximize the probability that $e_i$ and $e_j$ come from a positive pair, given $a_i = a_j$. However, it is different from $\mathcal{L}_{CE}$ in that it explicitly enforces pairwise constraints in the representation space $\phi(\cdot)$. During training, we jointly optimize $\mathcal{L}_{CE}$ and $\mathcal{L}_{CL}$:
\begin{align} \label{eqn:tot_loss}
    \mathcal{L} = \mathcal{L}_{CE} + \lambda \cdot \mathcal{L}_{CL},
\end{align}
where $\lambda$ is a balancing coefficient. This joint optimization, \method{}, improves upon $\mathcal{L}_{CE}$ by mining richer knowledge in the training data via encouraging meaningful pairwise relations in the representation space $\phi(\cdot)$. We conjecture that the model learns discriminative features in alignment with the classification objective. The effectiveness will be empirically examined (\secref{sec:experiments_human} and \secref{sect:results_nlg}) and qualitatively analyzed (\secref{sect:discuss_visualize}). 

\subsection{Implementation Details}

% \begin{table*}[t]
% \centering
% \begin{tabular}{c c c c c}
%   \toprule
%   Model & \blogsmall{} & \blogbig{} & \imdb{} & \enron{} \\
%   \midrule
%   Token SVM \citep{imdb62} & - & - & 92.5 & - \\
%   Char-CNN \citep{char_cnn} & 61.2 & 49.4 & 91.7 & - \\
%   Continuous N-gram \citep{conti_ngram} & 61.3 & 52.8 & 95.1 & - \\
%   N-gram CNN \citep{ngrams_cnn} & 63.7 & 53.1 & 95.2 & - \\
%   Syntax CNN \citep{syntax_cnn} & 64.1 & 56.7 & 96.2 & - \\
%   BertAA \citep{bertaa} & 65.4 & 59.7 & 93.0 & 97.1 \\ 
%   \midrule
%   \bertbase{} \textit{(our baseline)} & 60.4 & 55.2 & 97.2 & 97.6 \\
%   \bertcl{} & 66.3 (5.9\textcolor{black}{$\uparrow$}) & 62.0 (6.8\textcolor{black}{$\uparrow$}) & 97.9 (0.7\textcolor{black}{$\uparrow$}) & 97.6 \\
%   \debertabase{} \textit{(our baseline)} & 69.1 & 64.7 & 98.1 & 97.9 \\
%   \debertacl{} & \textbf{69.7 (0.6\textcolor{black}{$\uparrow$})} & \textbf{68.4 (3.7\textcolor{black}{$\uparrow$})} & \textbf{98.2 (0.1\textcolor{black}{$\uparrow$})} & \textbf{98.0 (0.1\textcolor{black}{$\uparrow$})} \\
%   \bottomrule
% \end{tabular}
% \caption{Results on human AA datasets (accuracy). Results in top section are from their respective papers. The best model for each dataset is \textbf{bolded}. Improvements over the baselines are indicated in parentheses. }
% \label{table:sota}
% \end{table*}

\begin{table*}[t]
\centering
\begin{tabular}{c c c c}
  \toprule
  \textbf{Model} & \textbf{\blogsmall{}} & \textbf{\blogbig{}} & \textbf{\imdb{}} \\
  \midrule
  Token SVM \citep{imdb62} & - & - & 92.5 \\
  Char-CNN \citep{char_cnn} & 61.2 & 49.4 & 91.7 \\
  Continuous N-gram \citep{conti_ngram} & 61.3 & 52.8 & 95.1 \\
  N-gram CNN \citep{ngrams_cnn} & 63.7 & 53.1 & 95.2  \\
  Syntax CNN \citep{syntax_cnn} & 64.1 & 56.7 & 96.2 \\
  BertAA \citep{bertaa} & 65.4 & 59.7 & 93.0  \\ 
  \midrule
  \bertbase{} \textit{(our baseline)} & 60.4 & 55.2 & 97.2  \\
  \bertcl{} & 66.3 (5.9\textcolor{black}{$\uparrow$}) & 62.0 (6.8\textcolor{black}{$\uparrow$}) & 97.9 (0.7\textcolor{black}{$\uparrow$}) \\
  \debertabase{} \textit{(our baseline)} & 69.1 & 64.7 & 98.1 \\
  \debertacl{} & \textbf{69.7 (0.6\textcolor{black}{$\uparrow$})} & \textbf{68.4 (3.7\textcolor{black}{$\uparrow$})} & \textbf{98.2 (0.1\textcolor{black}{$\uparrow$})}  \\
  \bottomrule
\end{tabular}
\caption{Results on human AA datasets, measured in accuracy.\footnote{Following previous work in this table, we specify a precision of 1 decimal place.} Results in top section are from their respective papers. Improvements over the baselines are indicated in parentheses. The best model for each dataset is \textbf{bolded}.}
\label{table:sota}
\end{table*}

We implement $\phi$ with two pre-trained transformer encoders, BERT \citep{bert} and DeBERTa \citep{deberta}. BERT is a commonly used text classification baseline and DeBERTa, its more recent counterpart. We use the \texttt{bert-base-cased} and \texttt{deberta-base} checkpoints from the \texttt{transformers} library \citep{huggingface}. For all datasets, the input length is set to $256$ and the embedding length per token is $768$. The transformer generates embeddings which are then passed to the classifier $h$. 

% We concatenate the token embeddings to obtain a sentence embedding of length $196608$. The embedding is then passed to the classifier $h$. 

We implement the classifier $h$ as a 2-layer Multi-Layer Perceptron (MLP) with a dropout of $0.35$. As described in \secref{sec:cl}, the final model $p$ is a composition of the pre-trained language model and the MLP classifier, \ie{} $p = \phi \circ h$.

In all experiments, we use the AdamW optimizer \citep{adamw} with an initial learning rate of $2e-5$ and a cosine learning rate schedule \citep{sgdr}. We train for $8$ epochs with a batch size of 24. We set $\lambda$ to 1.0 and $\tau$ to $0.1$. Training takes 2-12 hours depending on the dataset size with $4~\times$ RTX2080Ti. No model- or dataset-specific tuning was done for fair comparison and to show the robustness of the approach. 

\section{Human Authorship Attribution} \label{sec:experiments_human}

We first investigate the impact of contrastive learning on models for human authorship attribution.

\subsection{Experiment setup} 
\label{sec:human_exp_setup}

\paragraph{Models.} We experiment with four different models: two baselines \bertbase{} and \debertabase{}, fine-tuned with cross-entropy, and their \method{} versions, where X is the model name. These baselines allow us to isolate the effect of the proposed contrastive learning objective $\mathcal{L}_{CL}$. 

%We run each model on three popular authorship attribution corpora, which represent potential real-world use cases of identifying human authors.

%This is the standard task studied in most of the previous works. 

% We first conduct authorship attribution experiments on human-written materials. The experiment results show the effectiveness of 1) the fine-tuned pre-trained models and 2) the contrastive learning technique.  

% \subsection{Experimental Setup}

\paragraph{Datasets.} Following prior work \citep{char_cnn,syntax_cnn,bertaa}, we use the Blog \citep{blog} and IMDb \citep{imdb62} corpora for evaluation. For Blog, we take the top 10 and 50 authors with the most entries to form the \blogsmall{} and \blogbig{} datasets respectively. For IMDb, we take a standard subset of 62 authors \citep{imdb62} (\imdb). More details are in \appendixref{sect:statistics}.

\paragraph{Evaluation.} Following standard evaluation protocol, we divide each dataset into train/validation/test splits with an 8:1:1 ratio, and report the test split results here. Hyperparameter tuning, if any, is performed on the validation set. For easy comparison, we also present results on the 8:2 train/test splits used by \citet{bertaa} in \appendixref{sect:result82}. We do not observe any significant differences.

% We adopt the accepted machine learning best practice for evaluating our method. Instead of using 8:2 train/test splits as in \citet{bertaa}, we divide the dataset into train/validation/test splits with ratio 8:1:1. However, results on the 8:2 splits are also available in \appendixref{sect:result82}. The two sets of results differ only slightly.

% \subsection{Results}

% Full results are shown in \tabref{table:sota}.\footnote{The baseline results are reported by their respective papers, except for Continuous N-Gram \citep{conti_ngram} and N-gram CNN \citep{ngrams_cnn} which are reproduced and reported by \citet{syntax_cnn}.} 

\subsection{Results} 
From \tabref{table:sota}, we observe that the inclusion of contrastive learning improves the baseline performance across the board, allowing us to beat the previous state-of-the-art on all human AA datasets. We observe that the largest performance improvements come from \blogsmall{} and \blogbig{} datasets where there is substantial room for progress, \ie{} up to 6.8\% for BERT and 3.7\% for DeBERTa. In contrast, the performance gains on \imdb{} are marginal due to diminishing returns, with the baseline models already achieving close to 100\% accuracy. These results suggest that contrastive learning is empirically useful for fine-tuning pre-trained language models on the authorship attribution task, when the baseline performance is not approaching an asymptotic maximum.

% \textbf{\blogsmall{} and  \blogbig{}.}  Our approach with \bertcl{} improves the performance of \bertbase{} by 5.9\% and 6.8\% respectively. As \debertabase{} already achieves high performance on both datasets, the further improvements from \debertacl{} are relatively smaller. It is worth noting that \debertabase{} with simple finetuning already outperforms all previous results, showing the effectiveness of using learned language representations. \debertacl{} makes further gains on this baseline, showing the effect of the contrastive learning objective. 

% \textbf{\imdb{} and \enron{}.} Existing results on \imdb{} and \enron{} are already close to 100\%, making the margin for improvement small. On \imdb{}, both \bertcl{} and \debertacl{} make marginal gains on their non-contrastive counterparts. On \enron{}, \debertacl{} improves on \debertabase{} slightly.

% Our results show that learned language representations are effective for authorship attribution task, while contrastive learning further improves the performance. This empirically verifies the effectiveness of our strategies in \secref{sec:cl}.

\section{Synthetic Text Authorship Attribution} \label{sect:results_nlg}

% We investigate our proposed models on authorship attribution datasets with neural-generated text.

We investigate our proposed models on AA datasets with machine-generated text. This is to show how our method performs consistently across different dataset qualities and writers. Performing AA on human and machine authors together also reflects the increased importance of identifying machine-generated text sources. 

% In this section, we show that our method is general and can achieve state-of-the-art results on machine datasets. 

% our proposed models and loss function are proved useful in learning generated texts as well as differentiating human and machine writers. 

\subsection{Experimental Setup} \label{sect:results_nlg_setup}

\paragraph{Models.} We test the same four models from \secref{sec:experiments_human}: \bertbase{}, \bertcl{}, \debertabase{}, and \debertacl{}.

\paragraph{Dataset.} We use the \turingbench{} dataset \citep{turing_benchmark}. This corpus contains 200,000 news articles from 20 authors, \ie{} one human and 19 neural language generators (NLGs). The same set of article prompts is used for all authors to eliminate topical differences. The task objective is to attribute each text to one of the 20 writers. Note that this task implicitly encompasses the simpler binary classification task of machine text detection, where the 19 NLGs are treated as one machine writer. Additional dataset statistics are available in \appendixref{sect:statistics}. 

% This corpus contains 200,000 human-written or machine-generated texts from 20 sources, consisting of 1 human and 19 neural language generators. The 10,000 human-written texts are first collected from news articles from medias such as CNN, then 19 neural language models generate the rest 190,000 texts given the titles.  The motivation of the benchmark is that it resembles the \turingtest{} for neural language generation methods. 

%Thus, the topical difference between samples is smaller than previous humans datasets.

\paragraph{Evaluation.} We use the 7:1:2 train/validation/test splits provided by \citet{turing_benchmark} and report the results on the test set. 

\subsection{Results}

 \tabref{table:sota_nlg} shows the results of the synthetic authorship attribution benchmark.\footnote{Results of previous methods are from \turingbench{} \citep{turing_benchmark}. For consistency, we report results to 2 decimal places. For full results for other metrics, \ie{}  precision, recall, and F1-score, see \appendixref{sec:full_turing_results}.} Contrastive learning provides a small improvement in accuracy over the baseline models, in particular allowing \debertacl{} to set a new state-of-art. These results suggest that the use of general language representations and contrastive learning is generalizable to synthetic authorship attribution.
 
%  Our proposed model \debertacl{} outperformed all previous results, and further improves the performance of \debertabase{} by 0.53\%. Note that our \bertbase{} did not reach the previous reported result by BERT, which indicates that our method may not have reached its full performance given we did not perform hyperparameter tuning.

%Furthermore, it reveals that our approach may be applicable to relevant tasks such as fake review detection \citep{fake_review}.

\begin{table}[t]
\centering
\begin{tabular}{c c}
  \toprule
  \textbf{Model} & \textbf{\turingbench} \\
  \midrule
  Random Forest & 61.47 \\
  SVM (3-grams) & 72.99 \\
  WriteprintsRFC & 49.43 \\
  OpenAI Detector & 78.73 \\
%   \footnote{\url{https://huggingface.co/roberta-base-openai-detector}}
  Syntax CNN & 66.13 \\
  N-gram CNN & 69.14 \\
  N-gram LSTM-LSTM & 68.98 \\
  BertAA  & 78.12 \\ 
  BERT & 80.78 \\
  RoBERTa & 81.73 \\
  \midrule
  \bertbase{} \textit{(our baseline)} & 79.46 \\
  \bertcl{} & 80.59 (1.13\textcolor{black}{$\uparrow$})\\
  \debertabase{} \textit{(our baseline)}  & 82.00 \\
  \textbf{\debertacl{}} & \textbf{82.53 (0.53\textcolor{black}{$\uparrow$})}\\
  \bottomrule
\end{tabular}
\caption{Results on human and machine authorship attribution (accuracy). Results in the top section are from \citet{turing_benchmark}. Improvements over baselines are indicated in parentheses. Best model is \textbf{bolded}.}
\label{table:sota_nlg}
\end{table}

\section{Discussion} \label{sec:discussion}
In this section, we study the following questions:
\begin{itemize}[leftmargin=\parindent]\itemsep0em
    \item How does data availability affect the performance with and without contrastive learning?
    % \item How does availability of data affect the performance of fine-tuning? How can our approach influence the performance?
    \item How does contrastive learning qualitatively affect the representations learned?%Why does our proposed approach work? What representations does it learn?
    \item When does \method\ succeed and fail? 
\end{itemize}

% We will present corresponding analyzes in the following subsections. 

\subsection{Performance vs. Dataset Size}

\begin{figure}[h]
    \centering
    \includegraphics[width=1.0\columnwidth]{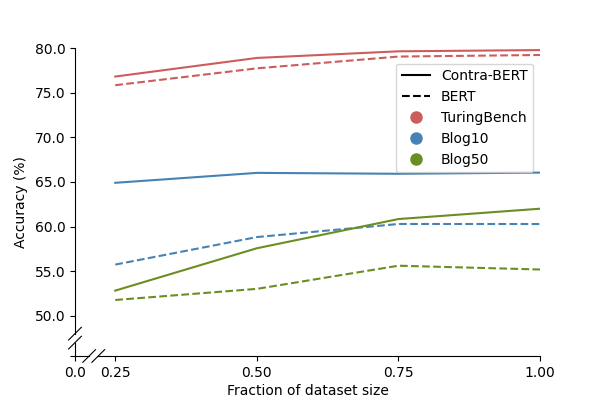}
    \caption{Comparison of performance between \bertbase{} and \bertcl{} under different data regimes. }
    \label{fig:small_dataset}
\end{figure}

% The availability of data has been a crucial concern for learning-based approaches. Thus, it is worthwhile to exam the impact of reducing the dataset size.

Due to the often-adversarial nature of real-world AA problems, the availability of appropriate data is a concern. Therefore, it is important to examine the impact of data availability on potential AA systems. To do this, we construct 4 subsets of the \blogsmall{}, \blogbig{}, and \turingbench{} datasets with stratified sampling by author. Each subset is 25\%, 50\%, 75\%, and 100\% the size of the original dataset. We use the same setup as in \secref{sec:human_exp_setup} to train \bertbase{} and \bertcl{} on each subset. 

% For example, a plagiarism-detection model is unlikely to have a large number of sample texts per student to draw from.

% Therefore, it is worth examining the impact of  To do this, we take stratified samples of the \blogsmall{} dataset based on authors. We shrink the dataset by 7 ratios, evenly spaced between $12.5\%$ and $87.5\%$ inclusive. Then use the same setup to train \bertbase{} and \bertcl{} on each subsampled set. The results are presented in \figref{fig:small_dataset}.

\figref{fig:small_dataset} plots accuracy vs. dataset size to illustrate the performance under different dataset sizes. On \blogsmall{}, \bertcl{} maintains a surprisingly consistent level of accuracy while \bertbase{} suffers significant degradation in performance as data decreases. On \blogbig{}, \bertcl{} shows more substantial performance gains compared to \bertbase{} as the dataset size increases. We hypothesize that the task is intrinsically harder due to the larger number of authors, requiring a larger amount of data to learn well. Even so, \method{} improves the performance of both BERT and DeBERTa by 6.8\% and 3.7\%, respectively, on the full dataset. On \turingbench{}, the difference in accuracy is less obvious, although \bertcl{} maintains the advantage. A possible explanation is that even the smaller subsets are sufficiently large. 

From the above statistics, we notice consistent improvements across different data regimes. A possible explanation is that the contrastive objective explicitly encourages the model to focus on inter-author differences as opposed to irrelevant features.

\begin{table*}[t]
    \centering
    \begin{tabular}{c | c c c c | c c}
      \toprule
      & \multicolumn{4}{c|}{\textbf{Feature Type}} & \multicolumn{2}{c}{\textbf{Performance Improvement (Acc.)}} \\
      Dataset & Content & Style & Hybrid & Topic & \bertbase\ & \debertabase\ \\
      \midrule
      \blogsmall{} & 0.82472 & \textbf{0.33766} & \textbf{0.59218} & 0.85465 & 5.9 & 0.6 \\
      \blogbig{} & 1.0000 & 1.0000 & 1.0000 & \textbf{0.81145} & 6.8 & 3.7\\
      \turingbench{} & \textbf{0.60842} & 0.56926 & 0.91988 & 1.0000 & 1.13 & 0.53 \\
    
    %   Content & 0.63299 & 0.76752 & \textbf{0.46698}\\
    %   Style & \textbf{0.000156} & 0.000462 & 0.000263\\
    %   Hybrid & \textbf{0.03363} & 0.05679 & 0.05224 \\
    %   Topic & 0.23303 & \textbf{0.22125} & 0.27266 \\
    %   \midrule
    %   \bertbase{} & 5.9 & \textbf{6.8} & 1.13 \\ 
    %   \debertabase{} & 0.6 & \textbf{3.7} & 0.53 \\ 
      \bottomrule
    \end{tabular}
    \caption{Inter-author difference on different feature metrics (improvements from each contrastive model listed for reference). The smaller the value, the higher the similarity measured by that feature. For consistency, each column is linearly scaled such that the maximum is 1. The smallest value for each feature is \textbf{bolded}.}
    \label{table:dataset_similarity_test}
\end{table*}

\subsection{Qualitative Representational Differences} \label{sect:discuss_visualize}

Next, we visualize the learned representations to understand the qualitative effect of the contrastive learning objective. We embed the test samples from the \blogbig{} dataset and visualize the result using t-SNE \citep{tsne}. 

Qualitatively, it is clear that \bertcl{} produces more distinct and tighter clusters compared to \bertbase{} (\figref{fig:tsne}). Since $\mathcal{L}_{CL}$ is the only independent variable in the experiment, differences in representation can be attributed to the contrastive objective. The improvement is expected, because the objective $\mathcal{L}_{CL}$ explicitly encourages the representation to be similar for intra-author samples (\ie{} tight clusters) and different for inter-author samples (\ie{} larger distance between clusters). This supports our conjecture in \secref{sec:cl}. 

% Qualitatively, this suggests that the contrastive loss objective is working as intended.

% From the confusion matrices, we can also observe performance improvements corresponding to changes in the visualization. Compared to \figref{fig:confusion}\subfig(c), the cells along the diagonal line in \figref{fig:confusion}\subfig(d) generally have a darker color, \ie{} higher class-level accuracy, and there are fewer dark points among the rest of the cells, \ie{} lower false positive and false negative rate. Overall, \bertcl{} has a $6.6\%$ improvement in absolute accuracy over \bertbase{}. We conjecture that the separation and concentration effects of $\mathcal{L}_{CL}$ contribute to this performance improvement. 

However, we observe that some clusters still overlap and are inseparable by t-SNE. This suggests that the model still faces some difficulty in distinguishing between specific authors.

\subsection{When Does \method{} Succeed and Fail?}

To understand the conditions in which Contra-X succeeds and fails, we follow \citet{topic_or_style} and extract $4$ stylometric features from the dataset: topic, style, content, and hybrid features. Detailed descriptions for each feature are in \appendixref{sec:similarity_metrics}. For this set of features, $\mathcal{F}$, the corresponding feature extractors are $\phi_f$, $f \in \mathcal{F}$. We can then represent each author, $A_i$, with a feature. Given an author $A_i$ with \textit{N} documents $\{t_i\}_{i=1:N}$, we define the representation of $A_i$ to be the mean of the vector representations of the $N$ documents: 

% \begin{equation} \label{avg_auth_feat}
%     \bar{F}_{feat}^{\mathcal{A}} = \frac{1}{n} \sum_{i}^{n} {F}_{feat}(t_i)
% \end{equation} 

\begin{equation} \label{avg_auth_feat}
    v_{A_i}^{f} = \frac{1}{N} \sum_{i=1}^N \phi_f (t_i). 
\end{equation} 

We analyze the relationship between model performance and dataset characteristics below. We exclude \imdb{} from this analysis since the maximum margin for improvement on the dataset is too small ($<3\%$). Performing analysis on these datasets may introduce confounding factors.

\paragraph{Dataset-level analysis.} %Our first objective is to look for pattern at the dataset level. To do this, we need to study the correlation between the features from a dataset and their respective margins of improvement from contrastive learning. 
Here, we wish to quantify the difficulty of distinguishing any two authors in each dataset and compare them against performance improvements. We define the inter-author dissimilarity of a dataset $\mathcal{D}$ in a feature space $f \in \mathcal{F}$ to be the mean pairwise difference across all author pairs $\langle A_i, A_j \rangle$ measured by the feature $f$:

% \begin{equation} \label{avg_dataset_feat}
%     \bar{F}_{feat}^{\textit{D}} = \frac{1}{n^{2}} \sum\limits_{\mathcal{A}_i \\ \in D}\sum\limits_{\mathcal{A}_j \in D} Dist(\bar{F}_{feat}^{\mathcal{A}_i}, \\ \bar{F}_{feat}^{\mathcal{A}_j})
% \end{equation} 

\begin{equation} \label{avg_dataset_feat}
    v_{\mathcal{D}}^f = \frac{1}{|A|^2}  \sum_{A_i, A_j \in \mathcal{D}} d(v_{A_i}^f, v_{A_j}^f), 
\end{equation} 

where $d$ is a distance metric for a pair of vectors:
% \vspace{-0.05in}
\begin{equation} 
d(v_{A_i}^f, v_{A_j}^f) = 
    \begin{cases}
    JSD(v_{A_i}^f, v_{A_j}^f) & \text{if} f = topic \\
    1 - \cos(v_{A_i}^f, v_{A_j}^f) & \text{otherwise}.
    \end{cases}
\end{equation} 

where JSD is the Jenson-Shannon Divergence \citep{jsd} and $\cos{}$ is the cosine similarity. The lower the value, the harder it is to distinguish the authors in a dataset in the corresponding feature space, on average. %The values for the 3 datasets are presented in \tabref{table:dataset_similarity_test}. 

% where $d$ is a distance metric for a pair of vectors. In our case, it is Jenson-Shannon Divergence (JSD) \citep{jsd} for $f = topic$, or (1 - cosine similarity) otherwise. 

From \tabref{table:dataset_similarity_test}, we observe that \blogbig{} has both the highest degree of topical similarity and the largest improvement from contrastive learning, while \turingbench{} has the least topical similarity and also the least improvement. This suggests that \method{} is robust to authors of similar topics. On the other hand, the opposite is true for content similarity: \turingbench{} has the highest content similarity and yet the least improvement.

% This may be explained by how content similarity as we define it measures word \textit{n}-gram frequencies, while our contrastive objective (\eqnref{eqn:cl_term}) is applied to the sentence embeddings $\phi(\cdot)$. This does not lead to significant information gain if the sentences contain similiar \textit{n}-grams.

% We notice that despite high style, hybrid, and topical similarity on \blogsmall{} and \blogbig{} datasets, we achieved significant performance gain. In contrast, \turingbench{} shows high content similarity but the performance improvement is relatively marginal. We hypothesize that our contrastive learning approach is more effective to distinguish similar texts interpreted by style, hybrid, and topical features. However, it might fall short when the content feature is similar. This is because our contrastive objective \eqnref{eqn:cl_term} is applied to the sentence embeddings $\phi(\cdot)$. This does not lead to significant information gain if the sentences are already similar measured by n-gram frequencies.

% \textbf{Does LDA or NLG work?} Latent Dirichlet Allocation (LDA) is often used to model topic distributions of human writings \citep{topic_or_style}. 

\textbf{Inadequacy of NLG models?} We also note the high topical dissimilarity of \turingbench{}. This is unexpected since this corpus is generated by querying each NLG model with the same set of titles as prompts (\secref{sect:results_nlg_setup}). Following \citet{topic_or_style}, we model topical similarity using Latent Dirichlet Allocation (LDA; \citealp{lda}). LDA represents a text as a distribution over latent topics, where each topic is represented as a distribution over words. This observation suggests that some NLG models may struggle to write on topic.\footnote{See \appendixref{sec:nlg_sample_analysis} for a brief analysis.} %although it is also possible that LDA is less suitable for modeling topical differences on neural-generated text.

% Here is another interesting observation. As described in \secref{sect:results_nlg_setup}, \turingbench{} is constructed by querying the machine authors to write about the same set of titles. This is to reduce the topical difference between authors \citep{turing_benchmark}. However, Latent Dirichlet Allocation (LDA) \citep{lda} reflects high topical dissimilarity on the dataset compared to the others. This either reflects the inadequacy of certain NLG models (\ie{} being unable to write on-topic) or suggests a room for improvement for topical models.

% a need for a different approach to identifying topics in future work.

% This suggests that when the authors in a dataset write about highly similar topics, the contrastive learning objective is particularly useful in finding minute difference between authors.

% In contrast, on the content metric, \turingbench{} shows the greatest similarity. We conjecture that this is correlated to the lack of performance gain from the contrastive models since content similarity measures word n-gram frequencies. Since our contrastive learning objective maximises the difference between learned representations at the word embedding level, a high degree of content similarity may predict the inability of the contrastive model to perform effectively.

% We note that there are multiple confounders present in this analysis, such as the number of authors in each dataset. Additionally, the feature metrics we use were designed for analysing human-authored text and may not fully capture the nuances of the machine-generated text in \turingbench{}.

\paragraph{Author-level analysis.} Next, we analyze how author characteristics affect the model performance on these authors. Specifically, we examine the correlation between the similarity of specific authors and how well the models distinguish between them. We define the distance between two authors to be the mean distance across all representation spaces:

% Going one step further, we wish to understand how the model performs on distinguishing specific authors in a dataset. Again, we conduct the study in a pairwise manner: we compute the similarity between a pair of authors and correlate it with the predictions that a model make on the authors. We define the distance of two authors as the mean distance across all representation spaces.

% \begin{equation} \label{pairwise_sim}
%     d(\mathcal{A}_i, \mathcal{A}_j) = \frac{1}{|F|}\sum_{feat} {\frac{1}{Norm} * Dist(\bar{F}_{feat}^{\mathcal{A}_i}, \\ \bar{F}_{feat}^{\bar{A}_j})}
% \end{equation} 

\begin{equation} \label{pairwise_sim}
    PD(A_i, A_j) = \frac{1}{|\mathcal{F}|} \sum_{f \in \mathcal{F}} \frac{1}{C_f} d(v_{A_i}^f, v_{A_j}^f),
\end{equation} 

where $C_f$ is a normalization term, defined as 

% where $Norm$ is the normalization term

% \begin{equation} \label{pairwise_sim}
%     Norm = {\max\limits_{m, n \in D}{(Dist(\bar{F}_{feat}^{\mathcal{A}_m}, \\ \bar{F}_{feat}^{\bar{A}_n}))}}
% \end{equation} 

% We conjecture that when two authors are highly similar, contrastive learning models can achieve improved class-level performance on those authors over a baseline model which may struggle to differentiate the two.

\begin{equation}
    C_f = \max_{A_i, A_j \in \mathcal{D}} d(v_{A_i}^f, v_{A_j}^f).
\end{equation}

\begin{figure}[t!]
    \small
    \centering
    \begin{subfigure}{\columnwidth}
        \includegraphics[width=\columnwidth]{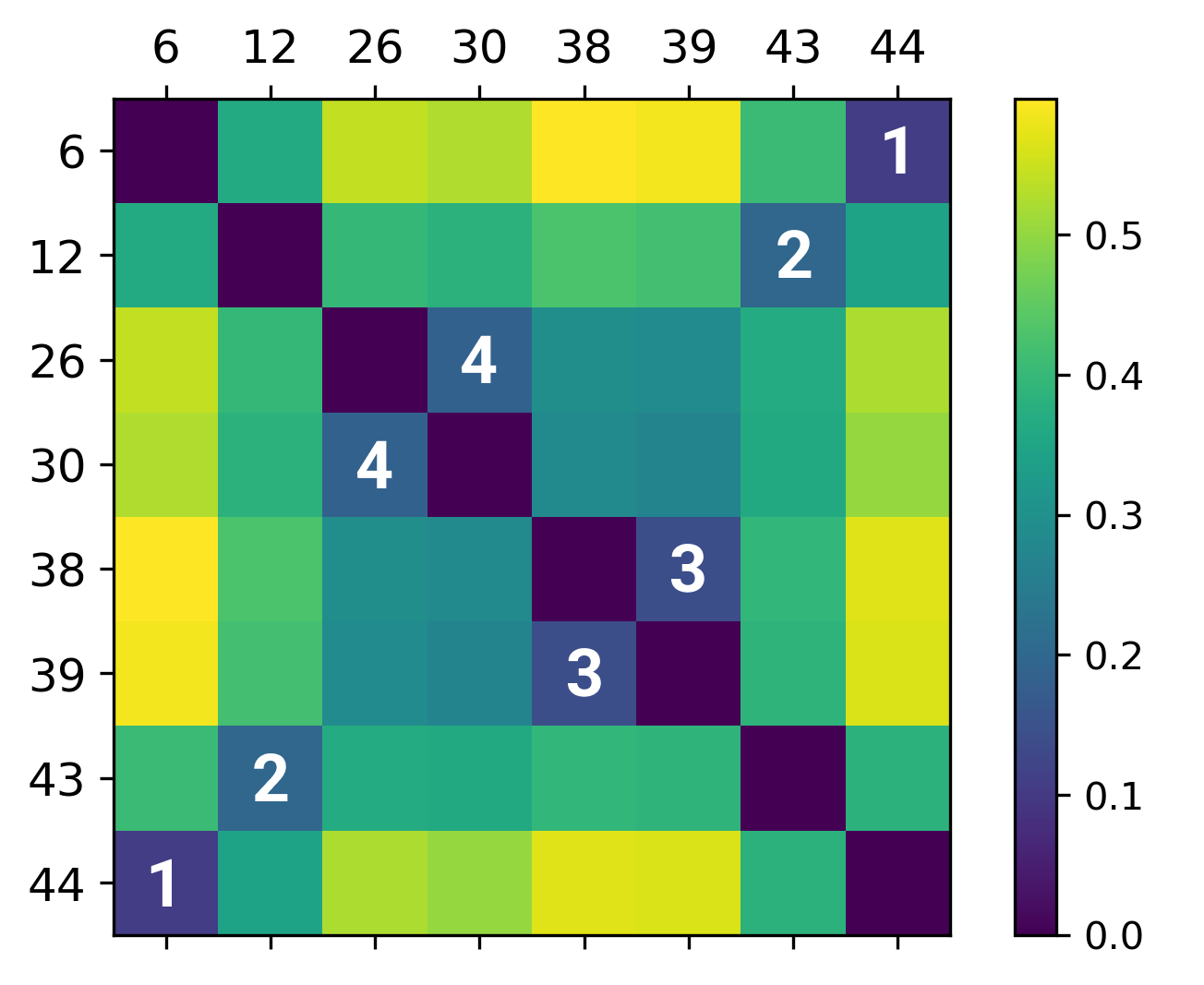}
        \caption{Feature dissimilarity matrix. Darker is more similar.}\label{fig:feature_matrix}
    \end{subfigure}
    \begin{subfigure}{\columnwidth}
    \includegraphics[width=\columnwidth]{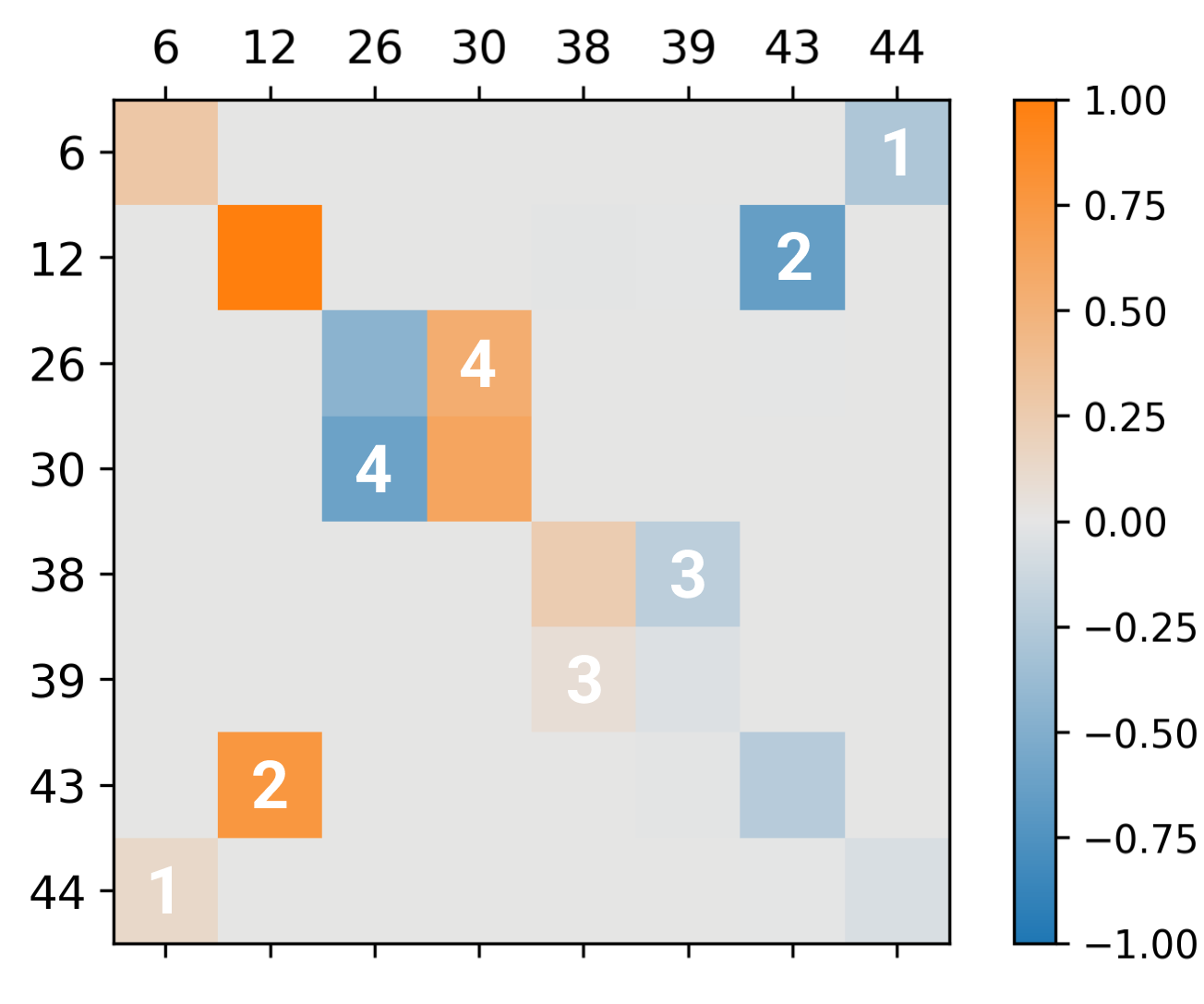}
    \caption{Relative confusion matrix. This is obtained by subtracting the confusion matrix of \bertbase{} from that of \bertcl{}.}\label{fig:confusion}
    \end{subfigure}
    \caption{Feature similarity matrix and relative confusion matrix between \bertbase{} and \bertcl{} on selected authors. In both figures, $(i, j)$ denotes the cell at the $i$-indexed row and $j$-indexed column. In (\subfig{a}), $(i, j)$ denotes $d(A_i, A_j)$, the feature dissimilarity between the two authors. In (\subfig{b}), a lower value (blue) of $(i, j)$ indicates \bertcl{} confused \Ai{} for \Aj{} less than \bertbase{}.}
    \label{fig:pair_cm_300}
\end{figure}

We plot the similarity matrix for selected \blogbig{} authors in \figref{fig:feature_matrix}. The authors are selected such that they form pairs that are highly indistinguishable by the above metrics. The cells numbered 1-4 represent the most similar author pairs (\ie{} darker-colored cells). Performance-wise, on each of these pairs, \bertcl{} shows significant improvements in overall class-level accuracy over \bertbase{}.\footnote{See \appendixref{sec:bert_similar_appendix} for exact values. This trend also holds for \debertacl{} and \debertabase{}; see \appendixref{sec:deberta_similar_appendix}.} This is consistent with the intuition that contrastive learning is more useful for distinguishing author pairs that are more similar.

% \figref{fig:pair_cm} shows selected authors from the \blogbig{} dataset. The left matrix shows the mean normalized similarities  across the four features $\bar{F}_{feat}^{\textit{A}}$ between authors (darker squares represent greater similarities). The squares numbered 1-4 represent the most similar author pairs.

\textbf{Increased bias.} The pairwise improvement mentioned above shows a curious property of being biased towards one of the authors in the pair. To visualize this, we subtract the confusion matrix of \bertbase{} from that of \bertcl{} and name the result the \textit{relative confusion matrix} (\figref{fig:confusion}). Each cell in the matrix indicates the increase in the probability that an author $A_i$ is classified as $A_j$ from \bertbase{} to \bertcl{}. For example, the blue cell at $(12, 43)$ shows that \bertcl{} confused $A_{12}$ as $A_{43}$ less than \bertbase{}, while the orange cell at $(43, 12)$ shows that \bertcl{} confused $A_{43}$ as $A_{12}$ more frequently.
%  the other. The colour of a given cell $(i, j)$, $i \neq j$, indicates whether \bertcl{} confused \Aj{} for \Ai{} more or less often than \bertbase{}. 

% which model confused \Aj{} for \Ai{} more often. For instance, the light-coloured $(43, 12)$ shows that \bertbase{} confused $A_{12}$ as $A_{43}$ more than \bertcl{}, while the darker $(12, 43)$ shows that \bertcl{} confused $A_{43}$ as $A_{12}$ more often.

% For instance, the light-coloured $(12, 43)$ shows that \bertbase{} misclassified 156 more samples of $A_{12}$ as $A_{43}$ than \bertcl{}, while the darker colour of $(43, 12)$ shows that \bertcl{} wrongly classified 131 more samples of $A_{43}$ as $A_{12}$.

Note first the intuitive link between the similarity and confusion matrices: similar authors are more likely to be confused by one of the models for each other. Observe also that the pairs in the confusion matrix are always present in light-dark pairs. In other words, if \bertbase{} misclassifies more samples from \Ai{} as \Aj{} (\eg{} $A_{12}$ as $A_{43}$), then \bertcl{} mislabels more samples from \Aj{} as \Ai{} (\ie{} $A_{43}$ as $A_{12}$). This suggests that as \bertcl{} learns to classify samples from \Ai{} better, it sacrifices the ability to identify \Aj{} samples. Note that although this sometimes stems from training on an imbalanced dataset, in our case, \Ai{} and \Aj{} contain similar numbers of samples.\footnote{See \appendixref{sec:bert_similar_appendix} for exact sample counts.} Thus, the observation is unlikely to be due to class imbalance. 

% The baseline models \bertbase{} and \debertabase{} systematically misclassify certain authors as other highly similar authors. With the addition of the contrastive learning objective, the models achieve a substantial increase in overall accuracy while systematically misclassifying a different set of authors. 

Nevertheless, the cumulative accuracy across \Ai{} and \Aj{} is always higher for \bertcl{} compared to the baseline, \eg{} 33.6\% vs 23.1\% for $A_{12}$ and $A_{43}$ combined, leading to an overall performance improvement on the whole dataset. This shows that the model implicitly learns to make trade-offs to optimize the contrastive objective, \ie{} it chooses to learn specialized representations that are particularly biased against some authors but improve the average performance over all authors. This shows that \method{} captures certain features that enable the model to distinguish a subset of the authors. However, to obtain consistent improvement, we need a deeper understanding of the difference between easily-confused authors and incorporate that insight into the contrastive learning algorithm \citep{no_free_lunch}. This can be potentially achieved by constructing more meaningful negative samples. However, this is beyond the scope of our paper and is left to future work.

\subsection{Potential Ethical Concerns}\label{sec:ethical}

In this subsection, we discuss potential ethical concerns related to the previous discussion on the increased bias in author-level performance.

\paragraph{Decreased fairness?} With classification models, fairness in predictions across classes is an important consideration. We want to, for instance, avoid demographic bias \citep{demographic_fairness}, which may manifest as systematic misclassifications of authors with specific sociolinguistic backgrounds.

% For instance, a model that predicts recidivism rates should not mispredict the outcome for some racial groups more than others \citep{recidivism}. In other words, it is undesirable for a model to show uneven performance across different classes.

Having observed increased bias against certain authors, we seek to find out if this trend holds across the entire dataset. We quantitatively evaluate this by computing the variance in class-level accuracy across all authors. The results show that the improvements from our contrastive learning objective appear to incur a penalty in between-author fairness. \bertcl{} on \blogsmall{} and \blogbig{}, and \debertacl{} on \blogbig{} achieve substantial gains in accuracy, and also produce notably higher variance than their baseline counterparts.\footnote{See \appendixref{sec:full_class_variance} for exact values.} In contrast, for models where the improvements are marginal, the differences in variance are insignificant. A potential direction for future work is investigating whether the use of contrastive learning consistently exacerbates variances in class-level accuracy. Studying the characteristics of the classes that the model is biased against may boost not just overall performance, but also predictive fairness.

\section{Conclusion} \label{sec:conclusion}
Successful authorship attribution necessitates the modeling of author-specific characteristics and idiosyncrasies. In this work, we made the first attempt to integrate contrastive learning with pre-trained language model fine-tuning on the authorship attribution task. We jointly optimized the contrastive objective and the cross-entropy loss, demonstrating improvements in performance on both human-written and machine-generated texts. We also showed our method is robust to dataset sizes and consistently improves upon cross-entropy fine-tuning under different data regimes. Critically, we contributed analyses of how and when Contra-X works in the context of the AA task. At the dataset level, we showed qualitatively that \method{} creates a tighter representation spread of each author and increased separation between authors. Within each dataset, at the author level, we found that Contra-X is able to distinguish between highly similar author pairs at the cost of hurting its performance on other authors. This points to a potential direction for future work, as resolving it would lead to better overall improvement and increased fairness of the final representation. 

\section*{Acknowledgments}
We thank Prof Min-Yen Kan for giving us comments on the project direction. We thank Ye Ma for insightful discussions. Bo and Yuchen are supported by NUS Science and Technology Scholarships. Samson was supported by Salesforce and Singapore’s Economic Development Board under its Industrial Postgraduate Programme. 

% The acknowledgments should go immediately before the references. Do not number the acknowledgments section.
% Do not include this section when submitting your paper for review.

\bibliography{references}

\begin{thebibliography}{43}
\expandafter\ifx\csname natexlab\endcsname\relax\def\natexlab#1{#1}\fi

\bibitem[{Bacciu et~al.(2019)Bacciu, La~Morgia, Mei, Nemmi, Neri, and
  Stefa}]{pan3}
Andrea Bacciu, Massimo La~Morgia, Alessandro Mei, Eugenio~Nerio Nemmi, Valerio
  Neri, and Julinda Stefa. 2019.
\newblock Cross-domain authorship attribution combining instance based and
  profile-based features.
\newblock In \emph{CLEF (Working Notes)}.

\bibitem[{Blei et~al.(2003)Blei, Ng, and Jordan}]{lda}
David~M. Blei, Andrew~Y. Ng, and Michael~I. Jordan. 2003.
\newblock \href {http://jmlr.org/papers/v3/blei03a.html} {Latent dirichlet
  allocation}.
\newblock \emph{J. Mach. Learn. Res.}, 3:993--1022.

\bibitem[{Brown et~al.(2020)Brown, Mann, Ryder, Subbiah, Kaplan, Dhariwal,
  Neelakantan, Shyam, Sastry, Askell, Agarwal, Herbert{-}Voss, Krueger,
  Henighan, Child, Ramesh, Ziegler, Wu, Winter, Hesse, Chen, Sigler, Litwin,
  Gray, Chess, Clark, Berner, McCandlish, Radford, Sutskever, and
  Amodei}]{gpt3}
Tom~B. Brown, Benjamin Mann, Nick Ryder, Melanie Subbiah, Jared Kaplan,
  Prafulla Dhariwal, Arvind Neelakantan, Pranav Shyam, Girish Sastry, Amanda
  Askell, Sandhini Agarwal, Ariel Herbert{-}Voss, Gretchen Krueger, Tom
  Henighan, Rewon Child, Aditya Ramesh, Daniel~M. Ziegler, Jeffrey Wu, Clemens
  Winter, Christopher Hesse, Mark Chen, Eric Sigler, Mateusz Litwin, Scott
  Gray, Benjamin Chess, Jack Clark, Christopher Berner, Sam McCandlish, Alec
  Radford, Ilya Sutskever, and Dario Amodei. 2020.
\newblock \href
  {https://proceedings.neurips.cc/paper/2020/hash/1457c0d6bfcb4967418bfb8ac142f64a-Abstract.html}
  {Language models are few-shot learners}.
\newblock In \emph{Advances in Neural Information Processing Systems 33: Annual
  Conference on Neural Information Processing Systems 2020, NeurIPS 2020,
  December 6-12, 2020, virtual}.

\bibitem[{Cust{\'o}dio and Paraboni(2018)}]{pan6}
Jos{\'e}~Eleandro Cust{\'o}dio and Ivandr{\'e} Paraboni. 2018.
\newblock Each-usp ensemble cross-domain authorship attribution.
\newblock \emph{Working Notes Papers of the CLEF}.

\bibitem[{Devlin et~al.(2018)Devlin, Chang, Lee, and Toutanova}]{bert}
Jacob Devlin, Ming{-}Wei Chang, Kenton Lee, and Kristina Toutanova. 2018.
\newblock \href {http://arxiv.org/abs/1810.04805} {{BERT:} pre-training of deep
  bidirectional transformers for language understanding}.
\newblock \emph{CoRR}, abs/1810.04805.

\bibitem[{Fabien et~al.(2020)Fabien, Villatoro{-}Tello, Motl{\'{\i}}cek, and
  Parida}]{bertaa}
Ma{\"{e}}l Fabien, Esa{\'{u}} Villatoro{-}Tello, Petr Motl{\'{\i}}cek, and
  Shantipriya Parida. 2020.
\newblock \href {https://aclanthology.org/2020.icon-main.16} {{BertAA} : {BERT}
  fine-tuning for authorship attribution}.
\newblock In \emph{Proceedings of the 17th International Conference on Natural
  Language Processing, {ICON} 2020, Indian Institute of Technology Patna,
  Patna, India, December 18-21, 2020}, pages 127--137. {NLP} Association of
  India {(NLPAI)}.

\bibitem[{Gao et~al.(2021)Gao, Yao, and Chen}]{simcse}
Tianyu Gao, Xingcheng Yao, and Danqi Chen. 2021.
\newblock \href {https://aclanthology.org/2021.emnlp-main.552} {Simcse: Simple
  contrastive learning of sentence embeddings}.
\newblock In \emph{Proceedings of the 2021 Conference on Empirical Methods in
  Natural Language Processing, {EMNLP} 2021, Virtual Event / Punta Cana,
  Dominican Republic, 7-11 November, 2021}, pages 6894--6910. Association for
  Computational Linguistics.

\bibitem[{Gollub et~al.(2013)Gollub, Potthast, Beyer, Busse, Rangel~Pardo,
  Rosso, Stamatatos, and Stein}]{forensics_plagiarism}
Tim Gollub, Martin Potthast, Anna Beyer, Matthias Busse, Francisco
  Rangel~Pardo, Paolo Rosso, Efstathios Stamatatos, and Benno Stein. 2013.
\newblock \href {https://doi.org/10.1007/978-3-642-40802-1_28} {Recent trends
  in digital text forensics and its evaluation}.
\newblock In \emph{CLEF}, volume 8138, pages 282--302.

\bibitem[{Gunel et~al.(2021)Gunel, Du, Conneau, and Stoyanov}]{supervised_cl}
Beliz Gunel, Jingfei Du, Alexis Conneau, and Veselin Stoyanov. 2021.
\newblock \href {https://openreview.net/forum?id=cu7IUiOhujH} {Supervised
  contrastive learning for pre-trained language model fine-tuning}.
\newblock In \emph{9th International Conference on Learning Representations,
  {ICLR} 2021, Virtual Event, Austria, May 3-7, 2021}. OpenReview.net.

\bibitem[{Gągała(2018)}]{pan5}
Łukasz Gągała. 2018.
\newblock Authorship attribution with neural networks and multiple features.
\newblock In \emph{Notebook for PAN at CLEF 2018}.

\bibitem[{Hardt et~al.(2016)Hardt, Price, and Srebro}]{demographic_fairness}
Moritz Hardt, Eric Price, and Nati Srebro. 2016.
\newblock \href
  {https://proceedings.neurips.cc/paper/2016/hash/9d2682367c3935defcb1f9e247a97c0d-Abstract.html}
  {Equality of opportunity in supervised learning}.
\newblock In \emph{Advances in Neural Information Processing Systems 29: Annual
  Conference on Neural Information Processing Systems 2016, December 5-10,
  2016, Barcelona, Spain}, pages 3315--3323.

\bibitem[{He et~al.(2021)He, Liu, Gao, and Chen}]{deberta}
Pengcheng He, Xiaodong Liu, Jianfeng Gao, and Weizhu Chen. 2021.
\newblock \href {https://openreview.net/forum?id=XPZIaotutsD} {Deberta:
  decoding-enhanced bert with disentangled attention}.
\newblock In \emph{9th International Conference on Learning Representations,
  {ICLR} 2021, Virtual Event, Austria, May 3-7, 2021}. OpenReview.net.

\bibitem[{Huang et~al.(2021)Huang, Ko, Tang, Liu, and
  Wu}]{supervised_cl_punctuation}
Qiushi Huang, Tom Ko, H.~Lilian Tang, Xubo Liu, and Bo~Wu. 2021.
\newblock \href {http://arxiv.org/abs/2107.09099} {Token-level supervised
  contrastive learning for punctuation restoration}.
\newblock \emph{CoRR}, abs/2107.09099.

\bibitem[{Iqbal et~al.(2010)Iqbal, Binsalleeh, Fung, and Debbabi}]{forensic_1}
Farkhund Iqbal, Hamad Binsalleeh, Benjamin C.~M. Fung, and Mourad Debbabi.
  2010.
\newblock \href {https://doi.org/10.1016/j.diin.2010.03.003} {Mining
  writeprints from anonymous e-mails for forensic investigation}.
\newblock \emph{Digit. Investig.}, 7(1-2):56--64.

\bibitem[{Jafariakinabad and Hua(2019)}]{style_aware}
Fereshteh Jafariakinabad and Kien~A. Hua. 2019.
\newblock \href {https://doi.org/10.1109/ICMLA.2019.00061} {Style-aware neural
  model with application in authorship attribution}.
\newblock In \emph{18th {IEEE} International Conference On Machine Learning And
  Applications, {ICMLA} 2019, Boca Raton, FL, USA, December 16-19, 2019}, pages
  325--328. {IEEE}.

\bibitem[{Jawahar et~al.(2020)Jawahar, Abdul-Mageed, and
  Lakshmanan}]{detection_survey}
Ganesh Jawahar, Muhammad Abdul-Mageed, and Laks Lakshmanan, V.S. 2020.
\newblock \href {https://doi.org/10.18653/v1/2020.coling-main.208} {Automatic
  detection of machine generated text: A critical survey}.
\newblock In \emph{Proceedings of the 28th International Conference on
  Computational Linguistics}, pages 2296--2309, Barcelona, Spain (Online).
  International Committee on Computational Linguistics.

\bibitem[{Kawakami et~al.(2020)Kawakami, Wang, Dyer, Blunsom, and van~den
  Oord}]{cl_speech}
Kazuya Kawakami, Luyu Wang, Chris Dyer, Phil Blunsom, and A{\"{a}}ron van~den
  Oord. 2020.
\newblock \href {https://doi.org/10.18653/v1/2020.findings-emnlp.106} {Learning
  robust and multilingual speech representations}.
\newblock In \emph{Findings of the Association for Computational Linguistics:
  {EMNLP} 2020, Online Event, 16-20 November 2020}, volume {EMNLP} 2020 of
  \emph{Findings of {ACL}}, pages 1182--1192. Association for Computational
  Linguistics.

\bibitem[{Kestemont et~al.(2019)Kestemont, Stamatatos, Manjavacas, Daelemans,
  Potthast, and Stein}]{pan1}
Mike Kestemont, Efstathios Stamatatos, Enrique Manjavacas, Walter Daelemans,
  Martin Potthast, and Benno Stein. 2019.
\newblock Overview of the cross-domain authorship attribution task at
  $\{$PAN$\}$ 2019.
\newblock In \emph{Working Notes of CLEF 2019-Conference and Labs of the
  Evaluation Forum, Lugano, Switzerland, September 9-12, 2019}, pages 1--15.

\bibitem[{Loshchilov and Hutter(2017)}]{sgdr}
Ilya Loshchilov and Frank Hutter. 2017.
\newblock \href {https://openreview.net/forum?id=Skq89Scxx} {{SGDR:} stochastic
  gradient descent with warm restarts}.
\newblock In \emph{5th International Conference on Learning Representations,
  {ICLR} 2017, Toulon, France, April 24-26, 2017, Conference Track
  Proceedings}. OpenReview.net.

\bibitem[{Loshchilov and Hutter(2019)}]{adamw}
Ilya Loshchilov and Frank Hutter. 2019.
\newblock \href {https://openreview.net/forum?id=Bkg6RiCqY7} {Decoupled weight
  decay regularization}.
\newblock In \emph{7th International Conference on Learning Representations,
  {ICLR} 2019, New Orleans, LA, USA, May 6-9, 2019}. OpenReview.net.

\bibitem[{Mendenhall(1887)}]{mendenhall1887characteristic}
Thomas~Corwin Mendenhall. 1887.
\newblock The characteristic curves of composition.
\newblock \emph{Science}, (214s):237--246.

\bibitem[{Nathanson(2013)}]{jsd}
Michael Nathanson. 2013.
\newblock \href {https://doi.org/10.4169/amer.math.monthly.120.02.182} {Review:
  Elements of information theory. john wiley and sons, inc., hoboken, nj, 2006,
  xxiv + 748 pp., {ISBN} 0-471-24195-4, {\textdollar}111.00. by thomas m. cover
  and joy a. thomas}.
\newblock \emph{Am. Math. Mon.}, 120(2):182--187.

\bibitem[{Rahgouy et~al.(2019{\natexlab{a}})Rahgouy, Giglou, Rahgooy,
  Sheykhlan, and Mohammadzadeh}]{rahgouy2019cross}
Mostafa Rahgouy, Hamed~Babaei Giglou, Taher Rahgooy, Mohammad~Karami Sheykhlan,
  and Erfan Mohammadzadeh. 2019{\natexlab{a}}.
\newblock Cross-domain authorship attribution: Author identification using a
  multi-aspect ensemble approach.
\newblock In \emph{CLEF (Working Notes)}.

\bibitem[{Rahgouy et~al.(2019{\natexlab{b}})Rahgouy, Giglou, Rahgooy,
  Sheykhlan, and Mohammadzadeh}]{pan2}
Mostafa Rahgouy, Hamed~Babaei Giglou, Taher Rahgooy, Mohammad~Karami Sheykhlan,
  and Erfan Mohammadzadeh. 2019{\natexlab{b}}.
\newblock Cross-domain authorship attribution: Author identification using a
  multi-aspect ensemble approach.
\newblock In \emph{CLEF (Working Notes)}.

\bibitem[{Ruder et~al.(2016)Ruder, Ghaffari, and Breslin}]{char_cnn}
Sebastian Ruder, Parsa Ghaffari, and John~G. Breslin. 2016.
\newblock \href {http://arxiv.org/abs/1609.06686} {Character-level and
  multi-channel convolutional neural networks for large-scale authorship
  attribution}.
\newblock \emph{CoRR}, abs/1609.06686.

\bibitem[{Sapkota et~al.(2015{\natexlab{a}})Sapkota, Bethard, Montes, and
  Solorio}]{char_ngram}
Upendra Sapkota, Steven Bethard, Manuel Montes, and Thamar Solorio.
  2015{\natexlab{a}}.
\newblock \href {https://doi.org/10.3115/v1/N15-1010} {Not all character
  n-grams are created equal: A study in authorship attribution}.
\newblock In \emph{Proceedings of the 2015 Conference of the North {A}merican
  Chapter of the Association for Computational Linguistics: Human Language
  Technologies}, pages 93--102, Denver, Colorado. Association for Computational
  Linguistics.

\bibitem[{Sapkota et~al.(2015{\natexlab{b}})Sapkota, Bethard,
  Montes{-}y{-}G{\'{o}}mez, and Solorio}]{style_ngram}
Upendra Sapkota, Steven Bethard, Manuel Montes{-}y{-}G{\'{o}}mez, and Thamar
  Solorio. 2015{\natexlab{b}}.
\newblock \href {https://doi.org/10.3115/v1/n15-1010} {Not all character
  n-grams are created equal: {A} study in authorship attribution}.
\newblock In \emph{{NAACL} {HLT} 2015, The 2015 Conference of the North
  American Chapter of the Association for Computational Linguistics: Human
  Language Technologies, Denver, Colorado, USA, May 31 - June 5, 2015}, pages
  93--102. The Association for Computational Linguistics.

\bibitem[{Sari et~al.(2018)Sari, Stevenson, and Vlachos}]{topic_or_style}
Yunita Sari, Mark Stevenson, and Andreas Vlachos. 2018.
\newblock \href {https://aclanthology.org/C18-1029/} {Topic or style? exploring
  the most useful features for authorship attribution}.
\newblock In \emph{Proceedings of the 27th International Conference on
  Computational Linguistics, {COLING} 2018, Santa Fe, New Mexico, USA, August
  20-26, 2018}, pages 343--353. Association for Computational Linguistics.

\bibitem[{Sari et~al.(2017)Sari, Vlachos, and Stevenson}]{conti_ngram}
Yunita Sari, Andreas Vlachos, and Mark Stevenson. 2017.
\newblock \href {https://doi.org/10.18653/v1/e17-2043} {Continuous n-gram
  representations for authorship attribution}.
\newblock In \emph{Proceedings of the 15th Conference of the European Chapter
  of the Association for Computational Linguistics, {EACL} 2017, Valencia,
  Spain, April 3-7, 2017, Volume 2: Short Papers}, pages 267--273. Association
  for Computational Linguistics.

\bibitem[{Schler et~al.(2006)Schler, Koppel, Argamon, and Pennebaker}]{blog}
Jonathan Schler, Moshe Koppel, Shlomo Argamon, and James~W. Pennebaker. 2006.
\newblock \href
  {http://www.aaai.org/Library/Symposia/Spring/2006/ss06-03-039.php} {Effects
  of age and gender on blogging}.
\newblock In \emph{Computational Approaches to Analyzing Weblogs, Papers from
  the 2006 {AAAI} Spring Symposium, Technical Report SS-06-03, Stanford,
  California, USA, March 27-29, 2006}, pages 199--205. {AAAI}.

\bibitem[{Seroussi et~al.(2014)Seroussi, Zukerman, and Bohnert}]{imdb62}
Yanir Seroussi, Ingrid Zukerman, and Fabian Bohnert. 2014.
\newblock \href {https://doi.org/10.1162/COLI\_a\_00173} {Authorship
  attribution with topic models}.
\newblock \emph{Comput. Linguistics}, 40(2):269--310.

\bibitem[{Shrestha et~al.(2017)Shrestha, Sierra, Gonz{\'a}lez, Montes, Rosso,
  and Solorio}]{ngrams_cnn}
Prasha Shrestha, Sebastian Sierra, Fabio Gonz{\'a}lez, Manuel Montes, Paolo
  Rosso, and Thamar Solorio. 2017.
\newblock \href {https://aclanthology.org/E17-2106} {Convolutional neural
  networks for authorship attribution of short texts}.
\newblock In \emph{Proceedings of the 15th Conference of the {E}uropean Chapter
  of the Association for Computational Linguistics: Volume 2, Short Papers},
  pages 669--674, Valencia, Spain. Association for Computational Linguistics.

\bibitem[{Solaiman et~al.(2019)Solaiman, Brundage, Clark, Askell,
  Herbert{-}Voss, Wu, Radford, and Wang}]{ntg_impact}
Irene Solaiman, Miles Brundage, Jack Clark, Amanda Askell, Ariel
  Herbert{-}Voss, Jeff Wu, Alec Radford, and Jasmine Wang. 2019.
\newblock \href {http://arxiv.org/abs/1908.09203} {Release strategies and the
  social impacts of language models}.
\newblock \emph{CoRR}, abs/1908.09203.

\bibitem[{Tang et~al.(2021)Tang, Nair, Wang, Wang, Desai, Wade, Li,
  Celikyilmaz, Mehdad, and Radev}]{cl_dialogue}
Xiangru Tang, Arjun Nair, Borui Wang, Bingyao Wang, Jai Desai, Aaron Wade,
  Haoran Li, Asli Celikyilmaz, Yashar Mehdad, and Dragomir Radev. 2021.
\newblock \href {http://arxiv.org/abs/2112.08713} {{CONFIT:} toward faithful
  dialogue summarization with linguistically-informed contrastive fine-tuning}.
\newblock \emph{CoRR}, abs/2112.08713.

\bibitem[{Tian et~al.(2020)Tian, Krishnan, and Isola}]{cl_multiview}
Yonglong Tian, Dilip Krishnan, and Phillip Isola. 2020.
\newblock \href {https://doi.org/10.1007/978-3-030-58621-8\_45} {Contrastive
  multiview coding}.
\newblock In \emph{Computer Vision - {ECCV} 2020 - 16th European Conference,
  Glasgow, UK, August 23-28, 2020, Proceedings, Part {XI}}, volume 12356 of
  \emph{Lecture Notes in Computer Science}, pages 776--794. Springer.

\bibitem[{Uchendu et~al.(2020)Uchendu, Le, Shu, and Lee}]{aa_nlg_2}
Adaku Uchendu, Thai Le, Kai Shu, and Dongwon Lee. 2020.
\newblock \href {https://doi.org/10.18653/v1/2020.emnlp-main.673} {Authorship
  attribution for neural text generation}.
\newblock In \emph{Proceedings of the 2020 Conference on Empirical Methods in
  Natural Language Processing (EMNLP)}, pages 8384--8395, Online. Association
  for Computational Linguistics.

\bibitem[{Uchendu et~al.(2021)Uchendu, Ma, Le, Zhang, and
  Lee}]{turing_benchmark}
Adaku Uchendu, Zeyu Ma, Thai Le, Rui Zhang, and Dongwon Lee. 2021.
\newblock \href {https://doi.org/10.18653/v1/2021.findings-emnlp.172}
  {{TURINGBENCH}: A benchmark environment for {T}uring test in the age of
  neural text generation}.
\newblock In \emph{Findings of the Association for Computational Linguistics:
  EMNLP 2021}, pages 2001--2016, Punta Cana, Dominican Republic. Association
  for Computational Linguistics.

\bibitem[{van~der Maaten and Hinton(2008)}]{tsne}
Laurens van~der Maaten and Geoffrey Hinton. 2008.
\newblock \href {http://jmlr.org/papers/v9/vandermaaten08a.html} {Visualizing
  data using t-sne}.
\newblock \emph{Journal of Machine Learning Research}, 9(86):2579--2605.

\bibitem[{Wolf et~al.(2019)Wolf, Debut, Sanh, Chaumond, Delangue, Moi, Cistac,
  Rault, Louf, Funtowicz, and Brew}]{huggingface}
Thomas Wolf, Lysandre Debut, Victor Sanh, Julien Chaumond, Clement Delangue,
  Anthony Moi, Pierric Cistac, Tim Rault, R{\'{e}}mi Louf, Morgan Funtowicz,
  and Jamie Brew. 2019.
\newblock \href {http://arxiv.org/abs/1910.03771} {Huggingface's transformers:
  State-of-the-art natural language processing}.
\newblock \emph{CoRR}, abs/1910.03771.

\bibitem[{Wolpert and Macready(1997)}]{no_free_lunch}
David~H. Wolpert and William~G. Macready. 1997.
\newblock \href {https://doi.org/10.1109/4235.585893} {No free lunch theorems
  for optimization}.
\newblock \emph{{IEEE} Trans. Evol. Comput.}, 1(1):67--82.

\bibitem[{Zellers et~al.(2019)Zellers, Holtzman, Rashkin, Bisk, Farhadi,
  Roesner, and Choi}]{grover}
Rowan Zellers, Ari Holtzman, Hannah Rashkin, Yonatan Bisk, Ali Farhadi,
  Franziska Roesner, and Yejin Choi. 2019.
\newblock \href
  {https://proceedings.neurips.cc/paper/2019/hash/3e9f0fc9b2f89e043bc6233994dfcf76-Abstract.html}
  {Defending against neural fake news}.
\newblock In \emph{Advances in Neural Information Processing Systems 32: Annual
  Conference on Neural Information Processing Systems 2019, NeurIPS 2019,
  December 8-14, 2019, Vancouver, BC, Canada}, pages 9051--9062.

\bibitem[{Zhang et~al.(2021)Zhang, Bui, Yoon, Chen, Liu, Xia, Tran, Chang, and
  Yu}]{contrastive_pt_ft}
Jianguo Zhang, Trung Bui, Seunghyun Yoon, Xiang Chen, Zhiwei Liu, Congying Xia,
  Quan~Hung Tran, Walter Chang, and Philip~S. Yu. 2021.
\newblock \href {https://doi.org/10.18653/v1/2021.emnlp-main.144} {Few-shot
  intent detection via contrastive pre-training and fine-tuning}.
\newblock In \emph{Proceedings of the 2021 Conference on Empirical Methods in
  Natural Language Processing, {EMNLP} 2021, Virtual Event / Punta Cana,
  Dominican Republic, 7-11 November, 2021}, pages 1906--1912. Association for
  Computational Linguistics.

\bibitem[{Zhang et~al.(2018)Zhang, Hu, Guo, and Mao}]{syntax_cnn}
Richong Zhang, Zhiyuan Hu, Hongyu Guo, and Yongyi Mao. 2018.
\newblock \href {https://doi.org/10.18653/v1/d18-1294} {Syntax encoding with
  application in authorship attribution}.
\newblock In \emph{Proceedings of the 2018 Conference on Empirical Methods in
  Natural Language Processing, Brussels, Belgium, October 31 - November 4,
  2018}, pages 2742--2753. Association for Computational Linguistics.

\end{thebibliography}
\bibliographystyle{acl_natbib}

\clearpage

\appendix
\label{sec:appendix}

\section{Dataset Statistics} \label{sect:statistics}
\tabref{table:statistics} presents statistics of the \blogsmall{}, \blogbig{}, \imdb{}, and \enron{} datasets.

\begin{table*}
\centering
\begin{tabular}{c c c c c c}
  \toprule
    & \blogsmall{} & \blogbig{} & \imdb{} & \turingbench\\
  \midrule
  \# authors & 10 & 50 & 62 & 20\\
  \# total documents & 23498 & 73275 & 61973 & 149561\\
  avg char / doc (no whitespace) & 407 & 439 & 1401 & 1063\\
  avg words / doc & 118 & 124 & 341 & 188\\
  \bottomrule
\end{tabular}
\caption{Statistics of the four datasets used in our experiments.}
\label{table:statistics}
\end{table*}

\begin{table*}[t]
\centering
\begin{tabular}{c c c c c}
  \toprule
  Model & \blogsmall{} & \blogbig{} & \imdb{} \\
  \midrule
  Token SVM \citep{imdb62} & - & - & 92.5 \\
  Char-CNN \citep{char_cnn} & 61.2 & 49.4 & 91.7 \\
  Continuous N-gram \citep{conti_ngram} & 61.3 & 52.8 & 95.1 \\
  N-gram CNN \citep{ngrams_cnn} & 63.7 & 53.1 & 95.2 \\
  Syntax CNN \citep{syntax_cnn} & 64.1 & 56.7 & \textbf{96.2} \\
  BertAA \citep{bertaa} & \textbf{65.4} & \textbf{59.7} & 93.0 & \\ 
  \midrule
  \bertbase{} & 60.3 & 55.6 & 97.2\\
  \bertcl{} & 66.0 (5.7\textcolor{black}{$\uparrow$}) & 62.2(6.6\textcolor{black}{$\uparrow$}) & 97.7(0.5\textcolor{black}{$\uparrow$})\\
  \debertabase{} & 68.0 & 65.0 & 98.1\\
  \textbf{\debertacl{}} & \textbf{69.9(1.9\textcolor{black}{$\uparrow$})} & \textbf{69.7(4.7\textcolor{black}{$\uparrow$})} & \textbf{98.2(0.1\textcolor{black}{$\uparrow$})}\\
  \bottomrule
\end{tabular}
\caption{Results of human authorship attribution - 8:2 train/test split}
\label{table:result_82}
\end{table*}

% \newpage

\section{Human Authorship Attribution Results with 8:2 Split} \label{sect:result82}
Following \citet{bertaa}, we divide the datasets into train-test splits at an 8:2 ratio for \blogsmall{}, \blogbig{}, and \imdb{} and follow the default split for \turingbench. We show the results on the test set in \tabref{table:result_82}.

\section{Similarity Metrics}
\label{sec:similarity_metrics}

Following \citet{topic_or_style}, we use four key metrics to analyze the characteristics of individual datasets (\ie{} samples written by a particular author, or all samples in a corpus). We describe these metrics in detail below:

\textbf{Content.} We measure the frequencies of the most common word unigrams, bigrams, and trigrams to produce a feature vector that represents an author's content preferences over each document.

\textbf{Style.} We combine multiple stylometric features, \ie{} average word length, number of short words, percentage of digits, percentage of upper-case letters, letter frequency, digit frequency, vocabulary richness, and frequencies of function words and punctuation, into a feature vector representing an author's writing style in a given document.

\textbf{Hybrid.} We measure the frequencies of the most common character bigrams and trigrams, to capture both content and style preferences of the author \citep{char_ngram} in a given document.

\textbf{Topic.} We use Latent Dirichlet Allocation (LDA) \citep{lda} to generate a probability distribution over an author's possible topics. We run LDA with 20 topics, as in \citet{topic_or_style}, and fit the data over 500 iterations.

\section{\turingbench{} Dataset Analysis}
\label{sec:nlg_sample_analysis}

Closer examination of the \turingbench{} dataset reveals that some models appear to produce fairly incoherent text. \tabref{table:nlg_samples} contains snippets from various models. Qualitatively, it is difficult to identify what the topic of each text is supposed to be; there appear to be multiple topics referenced in each text. This suggests that some of these models do not write on-topic, and consequently may explain why LDA reflects a high degree of topical dissimilarity between models.

On the other hand, at the phrase level, these models largely put out sensible phrases, \eg{} ``strong economic growth'', ``stunning game'', ``suspicious clicks''. We hypothesize that this is why the content similarity on \turingbench{} is comparatively higher, since the content metric measures word \textit{n}-gram frequencies.

\begin{table*}[t]
\centering
\begin{tabular}{p{0.25\linewidth} | p{0.7\linewidth}}
  \toprule
  \textbf{Model} & \textbf{Text} \\
  \midrule
  CTRL & ``apple gives tim cook \$384 million stock grant... steve jobs is set to receive an additional \$1.4 billion in cash... recovery needs but it also requires p le with skills not just on paper or through education training but, crucially, real work experience. those are two things which can only come if we have strong economic growth...''\\
  FAIR\_WMT19 & ``antoine helps real sociedad draw with valladolid... sociedad's goal in a 1-1 was highlight of stunning game played on night terrorist bombing attack manchester. tuesday, two bombs exploded central manchester arena during popular outdoor concert, killing 22 p le and injuring hundreds more..."\\
  GROVER\_MEGA & ``...the messages, which along message some will choose avoid draft, ready for qualification training are fake, according public affairs. do not respond spoof, requires suspicious clicks, pictures, or notes function, an official memo from issued thursday reads...''\\
  TRANSFORMER\_XL & ``carlos ghosn, mum on tokyo escape, unleashes a rambling defense of the state student-teacher training program in japan... as 2015, three universities (hiroshima, izumo, kawachi) accept all two degrees; they have also accepted each other. nevertheless, buddhist monks maintain that their colleges provide admission hindu traditions rather than admitting any religious instruction.''\\
  \midrule
\end{tabular}
\caption{Sample text snippets from various NLG models in the \turingbench{} dataset. }
\label{table:nlg_samples}
\end{table*}

% CTRL:
% "the messages, which along message some will choose avoid draft, ready for qualification training are fake, according public affairs. do not respond spoof, requires suspicious clicks, pictures, or notes function, an official memo from issued thursday reads. is false notification related government of united states america"

\section{Analysis of Similar Author Pairs}
\label{sec:similar_authors}

\subsection{\bertbase{} and \bertcl{}}
\label{sec:bert_similar_appendix}

\begin{figure*}[t!]
    \centering
    % \vspace*{-2in}
    \includegraphics[width=0.9\columnwidth]{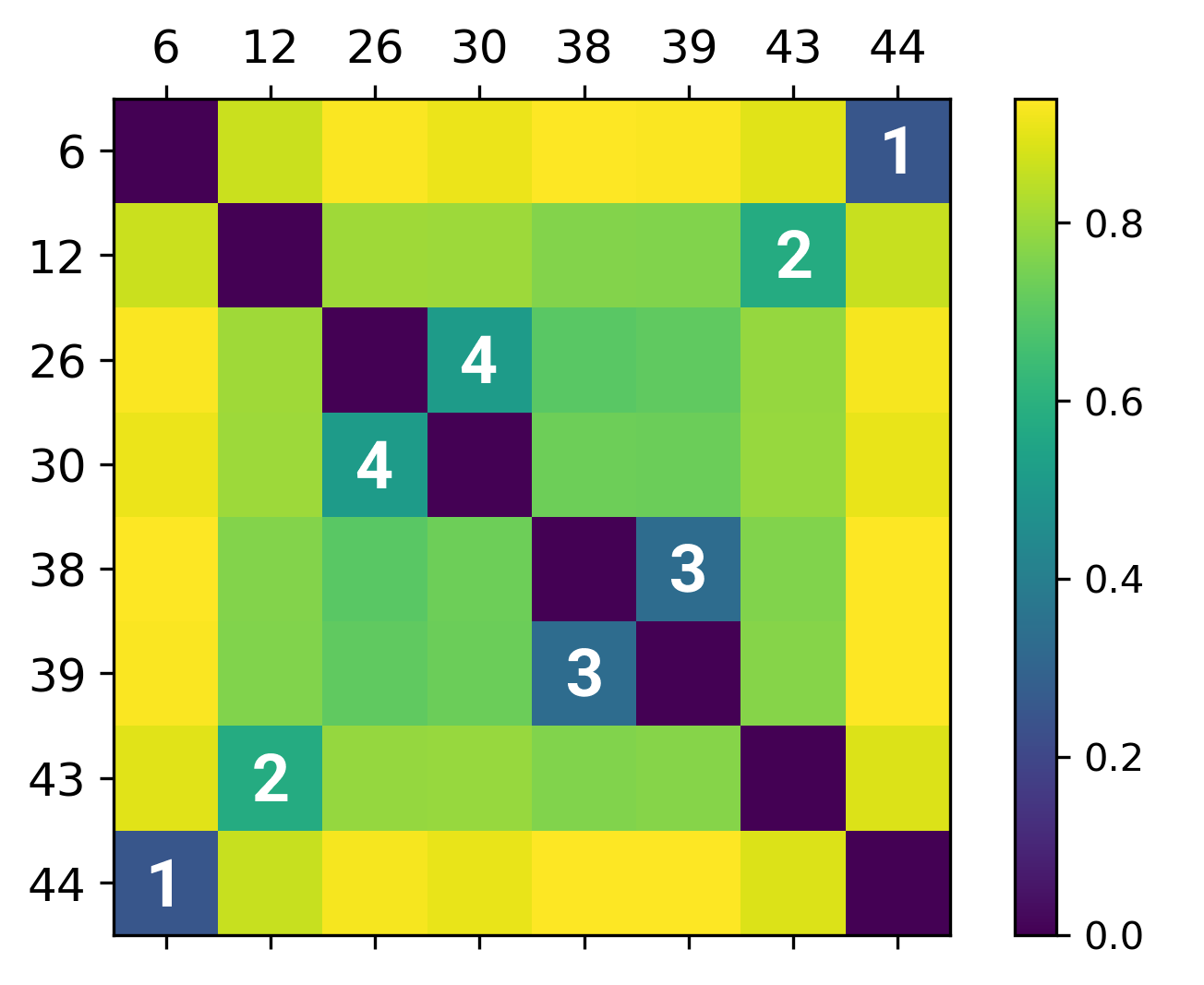}
    \includegraphics[width=0.9\columnwidth]{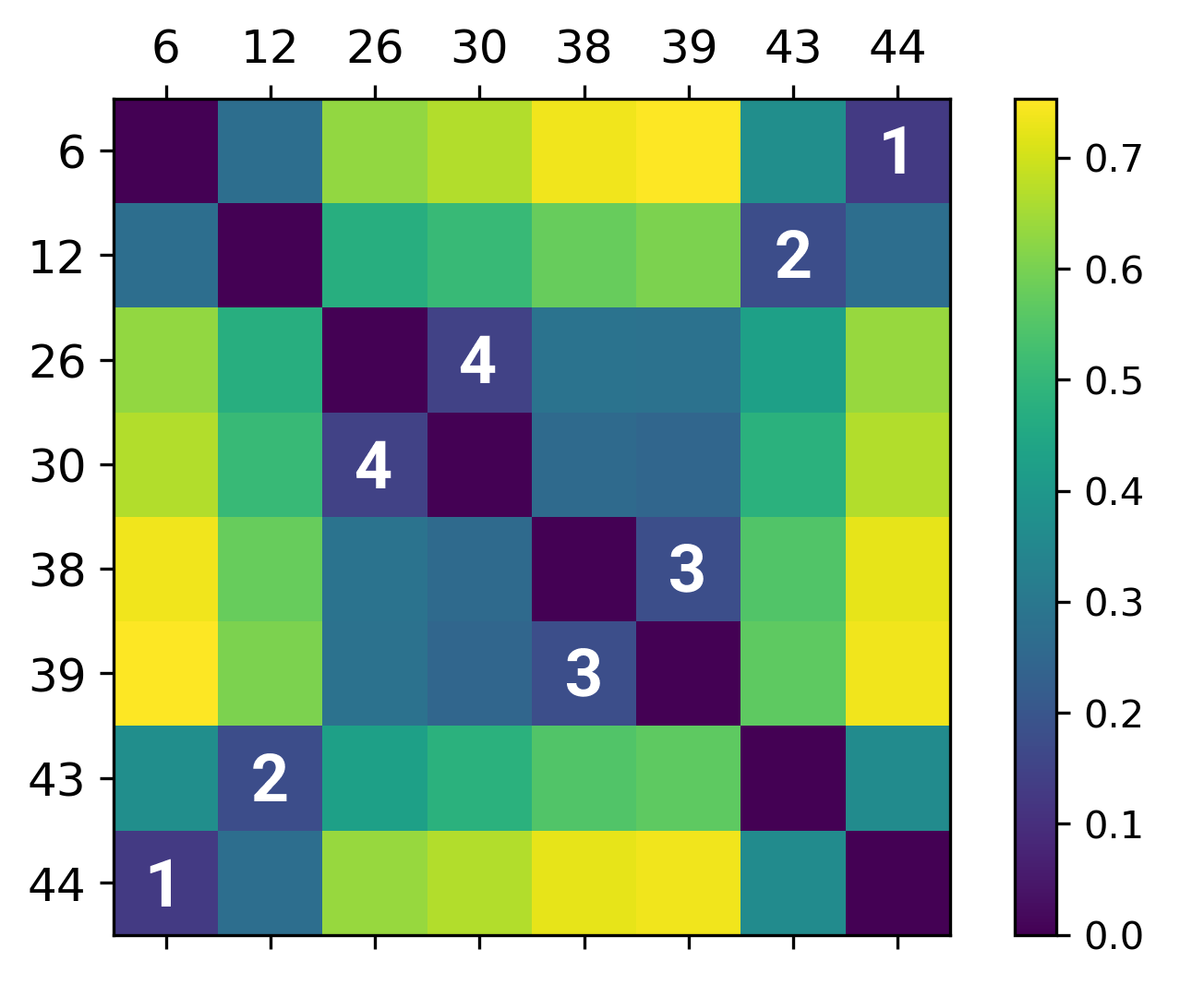}
    \includegraphics[width=0.9\columnwidth]{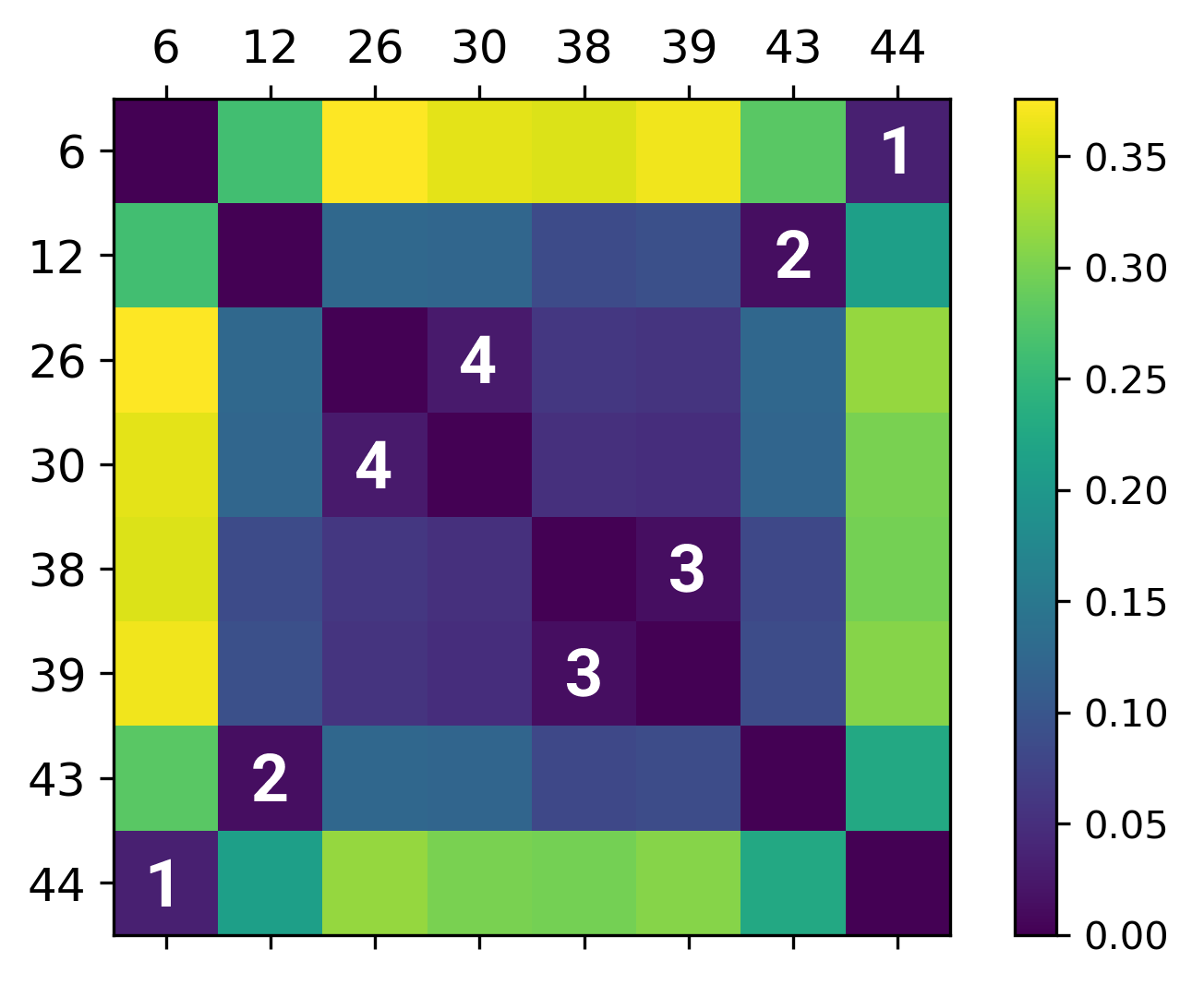}
    \includegraphics[width=0.9\columnwidth]{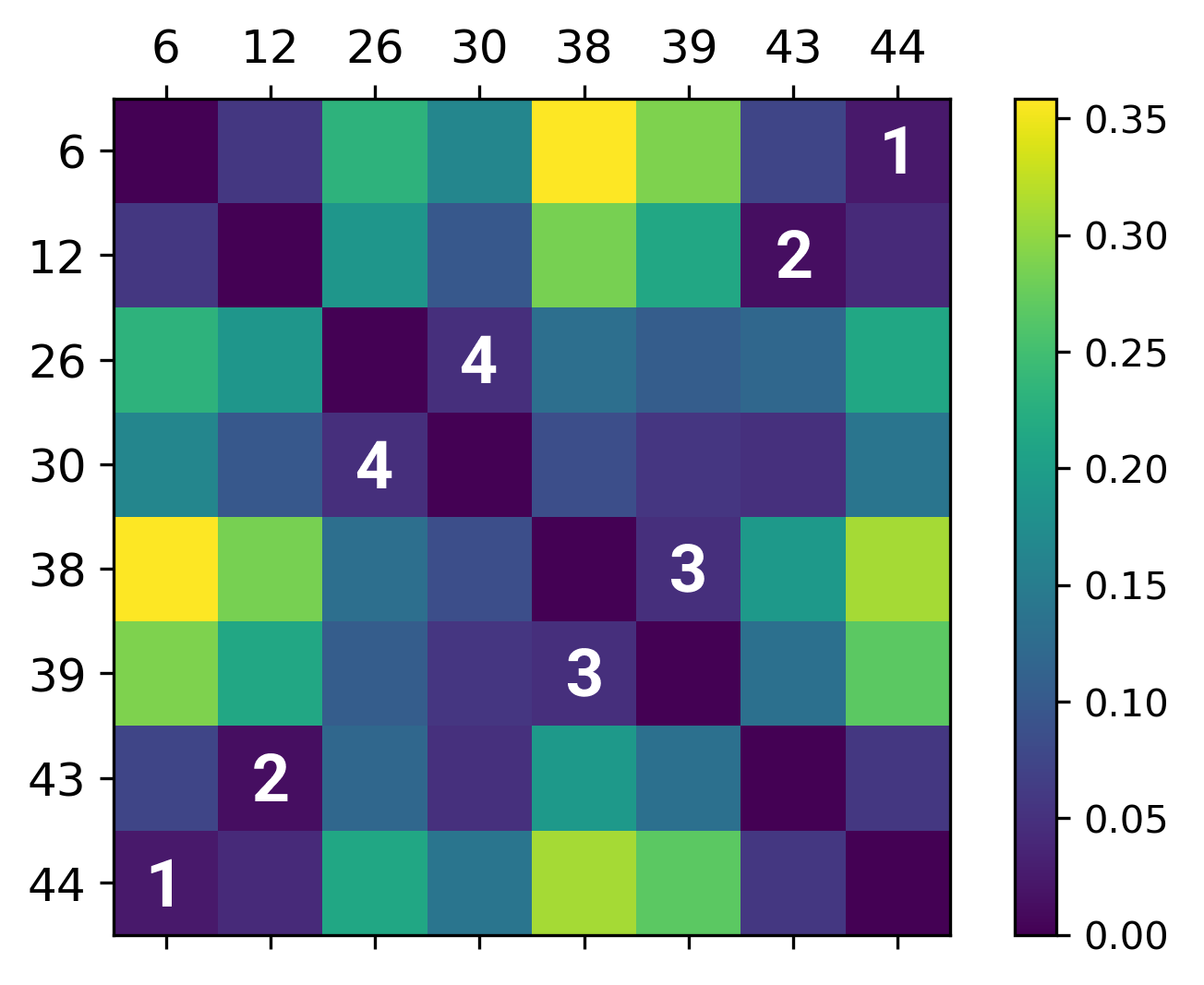}
    \caption{(Clockwise from top left) Similarity metrics between authors \Ai{} ($i$-indexed row) and \Aj{} ($j$-indexed column) for content, topic, hybrid, and style features respectively for selected authors on \blogbig{}.}
    \label{fig:pair_feat_cm_individual}
\end{figure*}

\figref{fig:pair_feat_cm_individual} shows the individual similarity matrices for the four feature types. The general pattern of the highlighted pairs being darker (\ie{} more similar) than their surrounding cells can be seen across all the matrices. \tabref{table:bert_blog50_pairs} shows the exact prediction accuracies for the four highlighted pairs. As noted previously, \bertcl{} always achieves a higher total accuracy (defined as total correct predictions over total samples) over both authors in a pair compared to \bertbase{}. 

\subsection{\debertabase{} and \debertacl{}}
\label{sec:deberta_similar_appendix}

\begin{figure*}[t]
    \centering\
    % \hspace*{-0.1in}
    \includegraphics[width=\columnwidth]{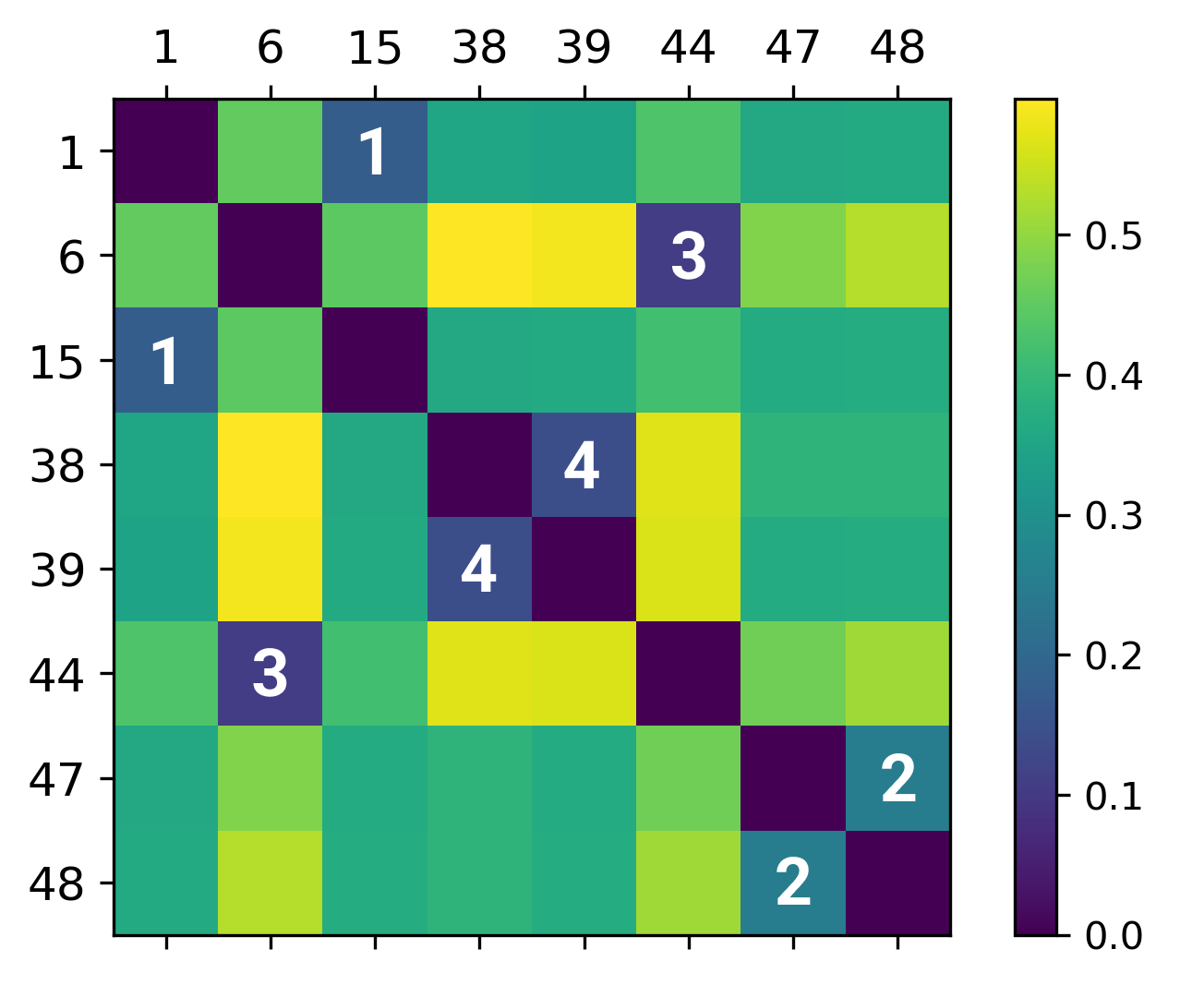}
    \includegraphics[width=\columnwidth]{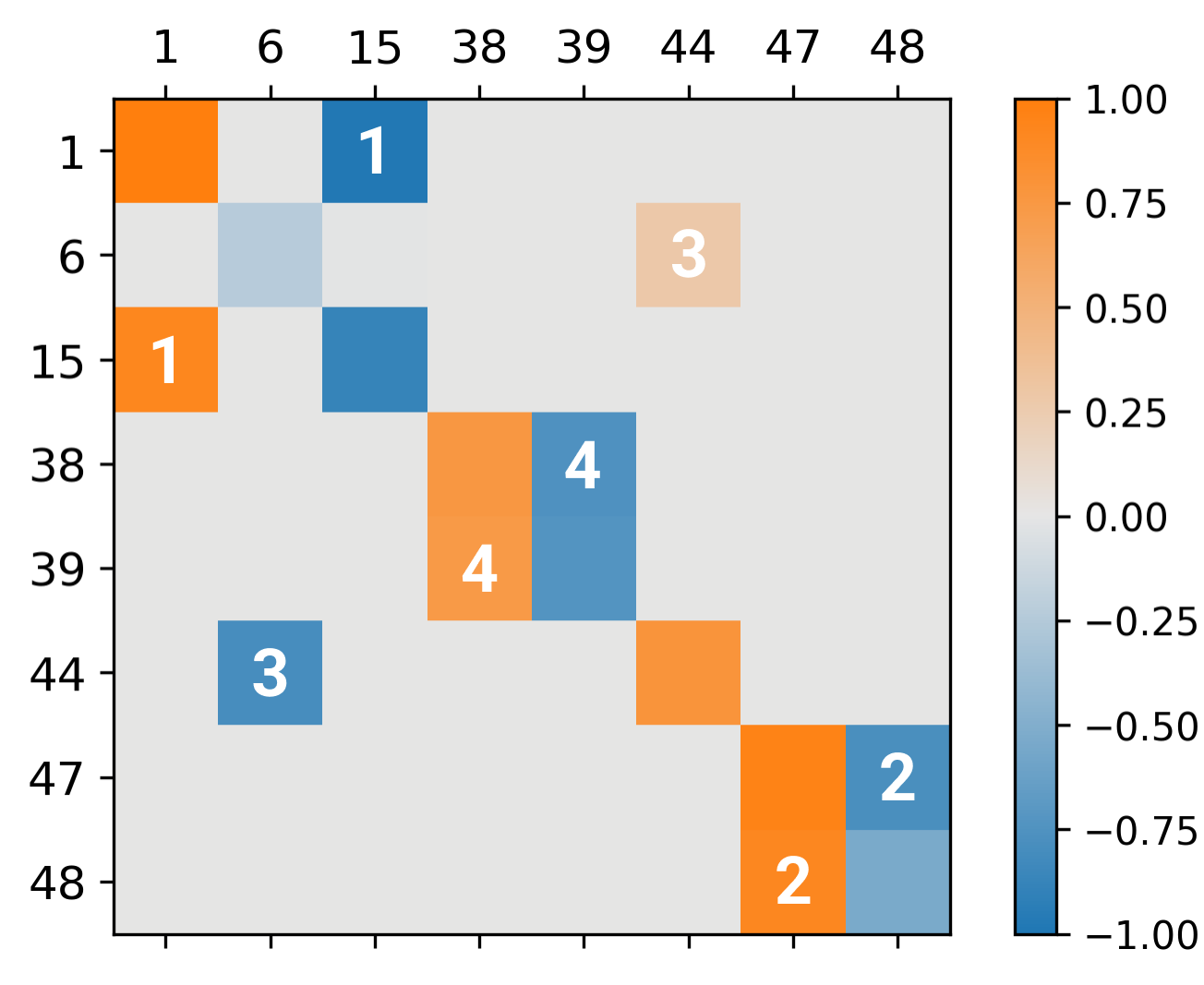}\\
    \small(\subfig{a}) Feature similarity matrix (left) and relative confusion matrix (right) between \debertabase{} and \debertacl{} on selected authors. For both figures, $(i, j)$ denotes the cell at the $i$-indexed row and $j$-indexed column. In the similarity matrix, $(i, j)$ denotes $d(A_i, A_j)$, the dissimilarity between the two authors (darker = more similar). In the confusion matrix, a lower value of $(i, j)$ indicates \debertacl{} confused \Ai{} for \Aj{} less than \debertabase{}.\\
    % $(i, j)$ denotes whether \debertacl{} confused \Ai{} for \Aj{} more (lighter) or less (darker) than \debertabase{}
    \vspace{0.5in}
    \includegraphics[width=\columnwidth]{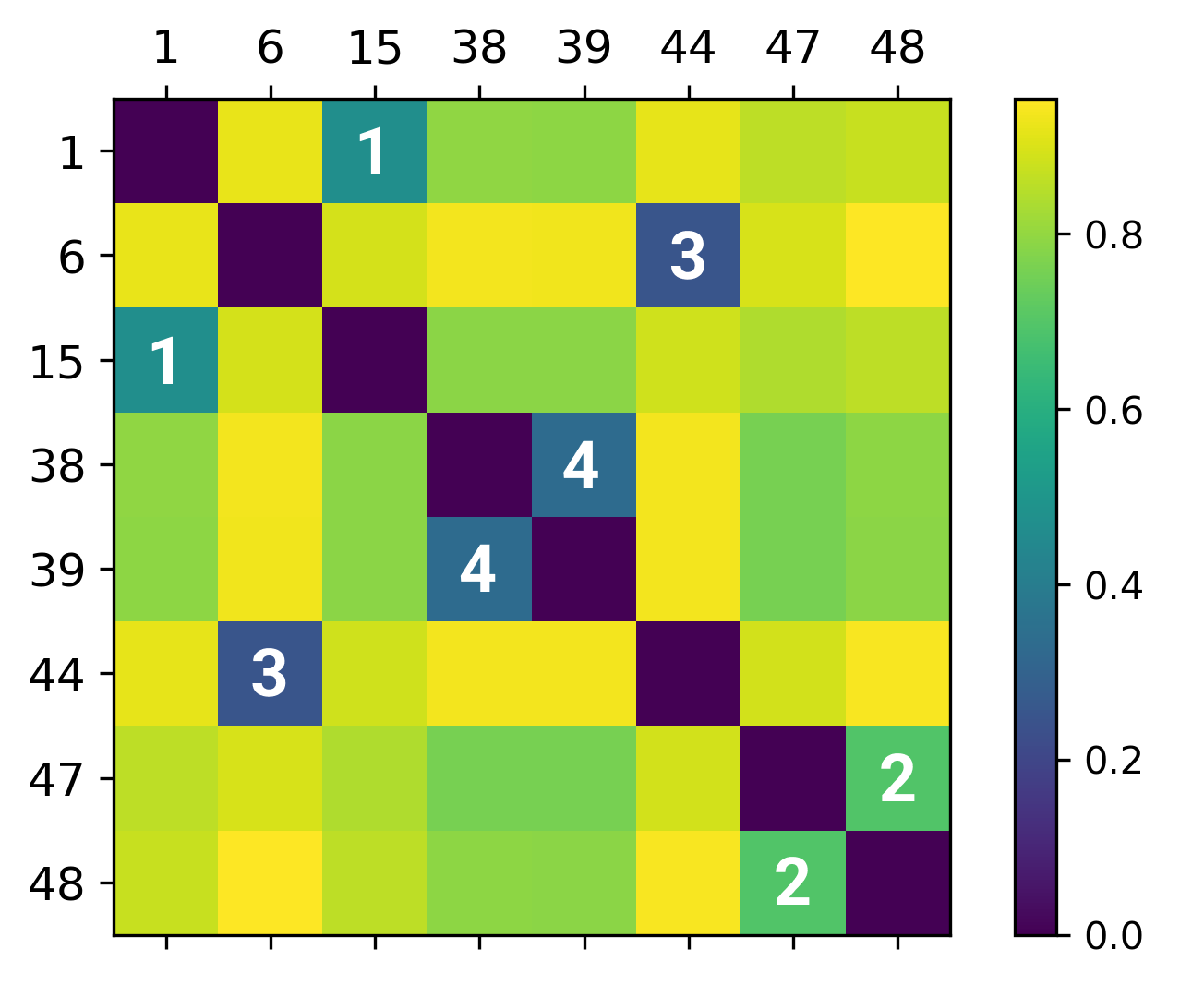}
    \includegraphics[width=\columnwidth]{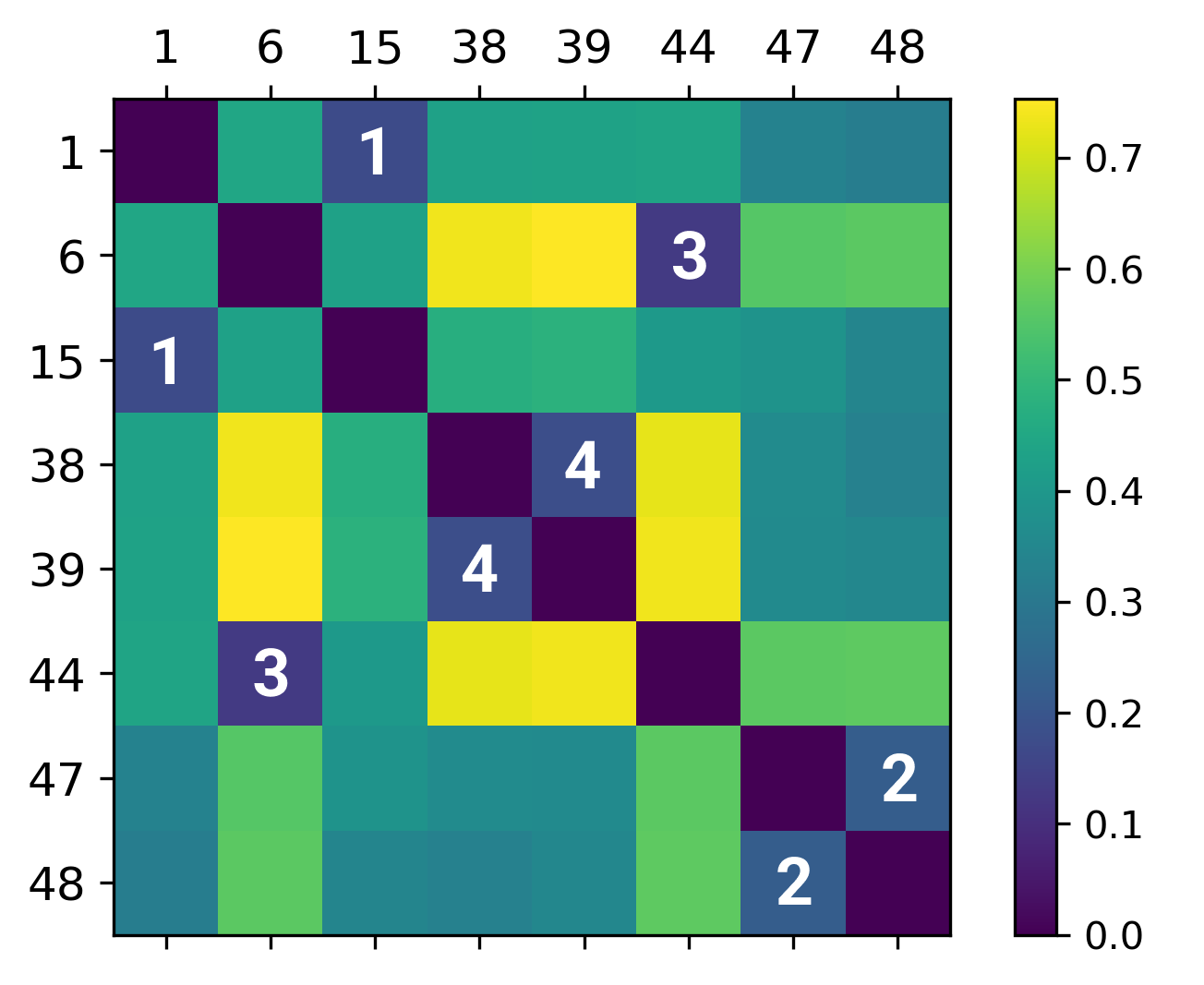}
    \includegraphics[width=\columnwidth]{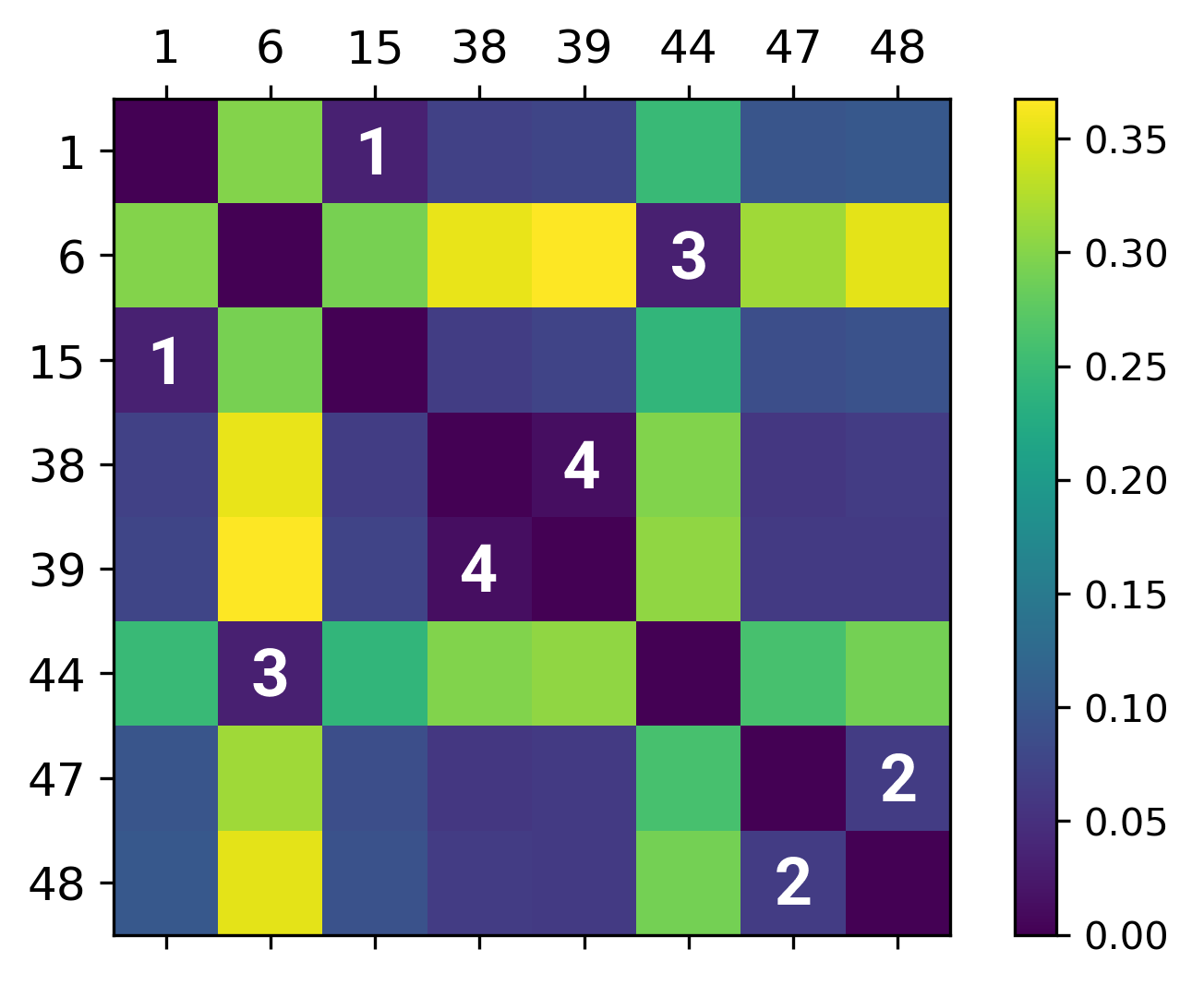}
    \includegraphics[width=\columnwidth]{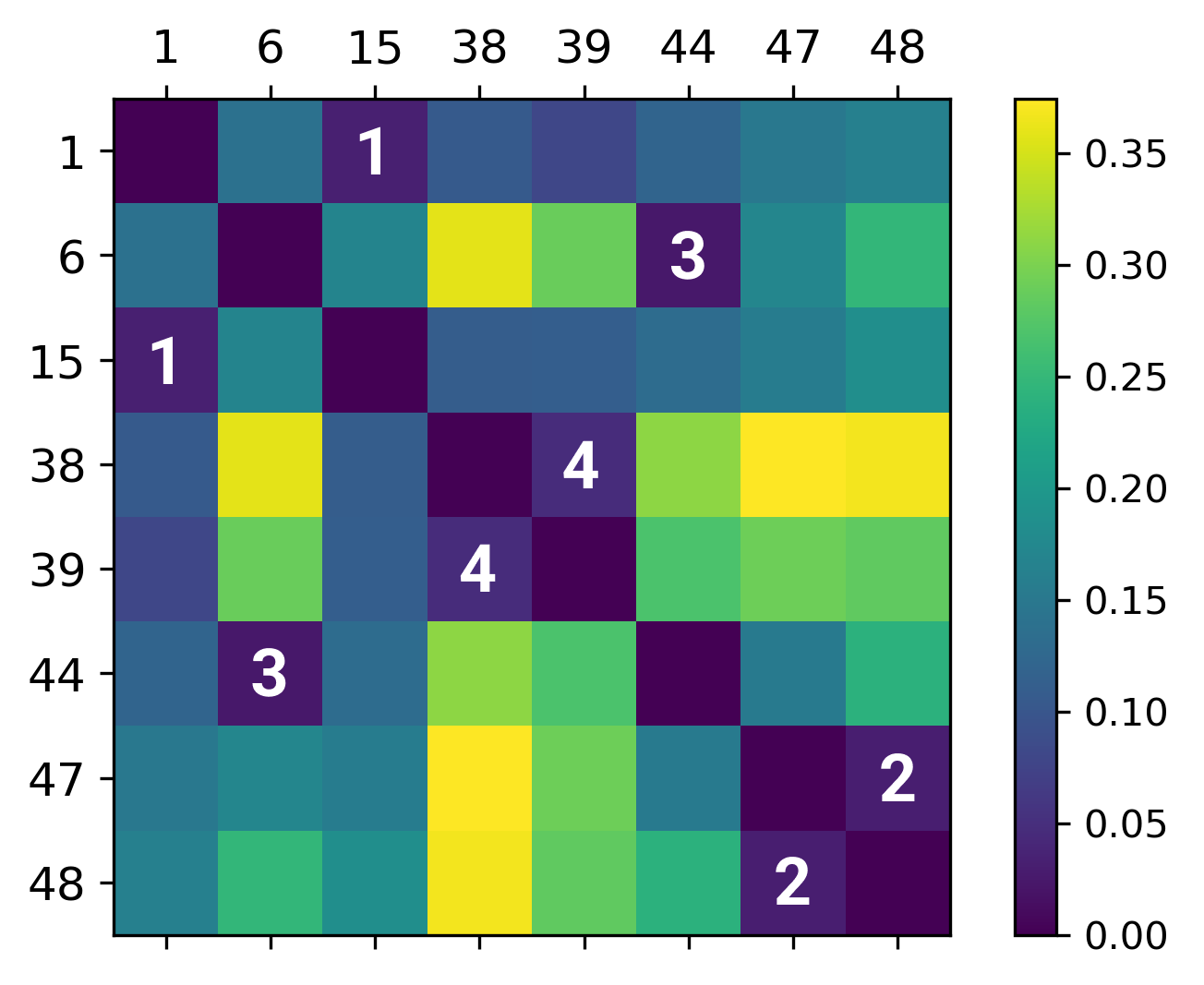}\\
    \small(\subfig{b}) (Clockwise from top left) Similarity metrics between authors \Ai{} ($i$-indexed row) and \Aj{} ($j$-indexed column) for content, topic, hybrid, and style features respectively for selected authors on \blogbig{}.\\
    \caption{Visualizations for selected author pairs for \debertabase{} and \debertacl{} on \blogbig{}.}
    \vspace*{0.5in}
    \label{fig:pair_2_cm}
\end{figure*}

\figref{fig:pair_2_cm} shows the feature similarity matrices and the relative confusion matrix for selected authors for \debertabase{} and \debertacl{}. Note that some of the author pairs are the same as those shown for \bertbase{} (\ie{} 6 \& 44, 38 \& 39) while other pairs are different. Similar to \figref{fig:pair_cm_300}(\subfig{b}), the colour of a given cell $(i, j)$, $i \neq j$, indicates whether \debertacl{} confused \Ai{} for \Aj{} more or less often than \debertabase{}. For instance, the blue-coloured $(1, 15)$ shows that \debertacl{} confused $A_{1}$ as $A_{15}$ less than \debertabase{}, while the orange $(15, 1)$ shows that \debertacl{} confused $A_{15}$ as $A_{1}$ more times.

\tabref{table:deberta_blog50_pairs} shows the exact prediction accuracies for the highlighted pairs. As with \bertcl{}, \debertacl{} achieves a higher total accuracy on each pair than \debertabase{}.

% \figref{fig:pair_2_feat_cm_individual} shows the individual similarity matrices for the four feature types.

\section{Full \turingbench{} results}
\label{sec:full_turing_results}
\tabref{table:sota_nlg_full} shows the precision, recall, F1, and accuracy scores on \turingbench{}.

\section{Class-Level Accuracy Variance}
\label{sec:full_class_variance}
\tabref{table:class_variance} shows the exact class-level accuracy variances for our four models on \blogsmall{}, \blogbig{}, and \turingbench{}.

\begin{table*}[t]
\centering
\begin{tabular}{c c c c c}
  \toprule
  \textbf{Model} & \textbf{Precision} & \textbf{Recall} & \textbf{F1} & \textbf{Accuracy} \\
  \midrule
  Random Forest & 58.93 & 60.53 & 58.47 & 61.47 \\
  SVM (3-grams) & 71.24 & 72.23 & 71.49 & 72.99 \\
  WriteprintsRFC & 45.78 & 48.51 & 46.51 & 49.43 \\
  OpenAI detector\footnote{\url{https://huggingface.co/roberta-base-openai-detector}} & 78.10 & 78.12 & 77.14 & 78.73 \\
  Syntax CNN & 65.20 & 65.44 & 64.80 & 66.13 \\
  N-gram CNN & 69.09 & 68.32 & 66.65 & 69.14 \\
  N-gram LSTM-LSTM & 6.694 & 68.24 & 66.46 & 68.98 \\
  BertAA & 77.96 & 77.50 & 77.58 & 78.12 \\ 
  BERT & 80.31 & 80.21 & 79.96 & 80.78 \\
  RoBERTa & 82.14 & 81.26 & 81.07 & 81.73 \\
  \midrule
  \bertbase{} \textit{(our baseline)} & 78.56 & 78.81 & 78.53 & 79.46 \\
  \bertcl{} & 80.10 (1.66\textcolor{black}{$\uparrow$}) & 79.99 (1.88\textcolor{black}{$\uparrow$}) & 79.84 (1.31\textcolor{black}{$\uparrow$}) & 80.59 (1.13\textcolor{black}{$\uparrow$})\\
  \debertabase{} \textit{(our baseline)} & 82.16 & 81.84 & 81.82 & 82.00 \\
  \textbf{\debertacl{}} & \textbf{82.84 (0.68\textcolor{black}{$\uparrow$})} & \textbf{82.04 (0.20\textcolor{black}{$\uparrow$})} & \textbf{81.98 (0.17\textcolor{black}{$\uparrow$})} & \textbf{82.53 (0.53\textcolor{black}{$\uparrow$})}\\
  \bottomrule
\end{tabular}
\caption{Full results across four metrics on human and machine authorship attribution. Results in the top section are from \citet{turing_benchmark}. Improvements over the baselines are indicated in parentheses. Best model is \textbf{bolded}. }
\label{table:sota_nlg_full}
\end{table*}

\newpage

\newpage
% \vspace*{-1in}
\begin{table*}[]
\centering
\vspace*{-1.35in}
\begin{tabular}{c | c c c | c c c | c }
  \hline
  & \multicolumn{3}{c|}{\textbf{Author 1}} & \multicolumn{3}{c|}{\textbf{Author 2}} & Total \\
  Model & \# & Samples & Correct & \# & Samples & Correct & Accuracy (\%) \\
  \hline \bertbase{} & \multirow{2}{*}{12} & \multirow{2}{*}{229} & 2 & \multirow{2}{*}{43} & \multirow{2}{*}{225} & 47 & 10.8 \\
  \bertcl{} & & & 209 & & & 0 & \textbf{46.0} \\
  \hline \bertbase{} & \multirow{2}{*}{30} & \multirow{2}{*}{153} & 8 & \multirow{2}{*}{26} & \multirow{2}{*}{154} & 92 & 32.6 \\
  \bertcl{} & & & 135 & & & 0 & \textbf{44.0} \\
  \hline \bertbase{} & \multirow{2}{*}{6} & \multirow{2}{*}{116} & 35 & \multirow{2}{*}{44} & \multirow{2}{*}{113} & 18 & 23.1 \\
  \bertcl{} & & & 73 & & & 4 & \textbf{33.6} \\
  \hline \bertbase{} & \multirow{2}{*}{38} & \multirow{2}{*}{112} & 48 & \multirow{2}{*}{39} & \multirow{2}{*}{112} & 8 & 25.0 \\
  \bertcl{} & & & 96 & & & 0 & \textbf{42.9} \\
  %\bertcl{} & 12 & 229 & 209 & 43 & 225 & 0 & \textbf{46.0} \\
%   \hline \bertbase{} & 30 & 153 & 8 & 26 & 154 & 92 & 32.6 \\
%   \bertcl{} & 30 & 153 & 135 & 26 & 154 & 0 & \textbf{44.0} \\
%   \hline \bertbase{} & 6 & 116 & 35 & 44 & 113 & 18 & 23.1 \\
%   \bertcl{} & 6 & 116 & 73 & 44 & 113 & 4 & \textbf{33.6} \\
%   \hline \bertbase{} & 38 & 112 & 48 & 39 & 112 & 8 & 25.0 \\
%   \bertcl{} & 38 & 112 & 96 & 39 & 112 & 0 & \textbf{42.9} \\
  \hline
\end{tabular}
\caption{Performance of \bertbase{} and \bertcl{} on selected author pairs of \blogbig{}. Higher accuracy for each pair is \textbf{bolded}.}
% \vspace*{1in}
\label{table:bert_blog50_pairs}
\end{table*}

\newpage
\begin{table*}
\centering
\vspace*{-2.5in}
\begin{tabular}{c | c c c | c c c | c }
  \hline
  & \multicolumn{3}{c|}{\textbf{Author 1}} & \multicolumn{3}{c|}{\textbf{Author 2}} & Total \\
  Model & \# & Samples & Correct & \# & Samples & Correct & Accuracy (\%) \\
  \hline \debertabase{} & \multirow{2}{*}{1} & \multirow{2}{*}{109} & 0 & \multirow{2}{*}{15} & \multirow{2}{*}{103} & 94 & 44.3 \\
  \debertacl{} & & & 107 & & & 0 & \textbf{50.5} \\
  \hline \debertabase{} & \multirow{2}{*}{47} & \multirow{2}{*}{105} & 0 & \multirow{2}{*}{48} & \multirow{2}{*}{104} & 61 & 29.2 \\
  \debertacl{} & & & 102 & & & 4 & \textbf{50.7} \\
  \hline \debertabase{} & \multirow{2}{*}{44} & \multirow{2}{*}{113} & 24 & \multirow{2}{*}{6} & \multirow{2}{*}{116} & 28 & 22.7 \\
  \debertacl{} & & & 108 & & & 3 & \textbf{48.5} \\
  \hline \debertabase{} & \multirow{2}{*}{38} & \multirow{2}{*}{112} & 0 & \multirow{2}{*}{39} & \multirow{2}{*}{112} & 90 & 40.2 \\
  \debertacl{} & & & 81 & & & 12 & \textbf{41.5} \\
  %\bertcl{} & 12 & 229 & 209 & 43 & 225 & 0 & \textbf{46.0} \\
%   \hline \bertbase{} & 30 & 153 & 8 & 26 & 154 & 92 & 32.6 \\
%   \bertcl{} & 30 & 153 & 135 & 26 & 154 & 0 & \textbf{44.0} \\
%   \hline \bertbase{} & 6 & 116 & 35 & 44 & 113 & 18 & 23.1 \\
%   \bertcl{} & 6 & 116 & 73 & 44 & 113 & 4 & \textbf{33.6} \\
%   \hline \bertbase{} & 38 & 112 & 48 & 39 & 112 & 8 & 25.0 \\
%   \bertcl{} & 38 & 112 & 96 & 39 & 112 & 0 & \textbf{42.9} \\
  \hline
\end{tabular}
\caption{Performance of \debertabase{} and \debertacl{} on selected author pairs of \blogbig{}. Higher accuracy for each pair is \textbf{bolded}.} 
% \vspace*{4in}
\label{table:deberta_blog50_pairs}
\end{table*}

\begin{table*}[t]
    \centering
    \vspace*{-2.5in}
    \begin{tabular}{c c c c}
    \toprule
     & \textbf{\blogsmall{}} & \textbf{\blogbig{}} & \textbf{\turingbench{}} \\
    \midrule
    \bertbase{} & 0.15494 & 0.10430 & 0.06747 \\
    \bertcl{} & \textbf{0.17698} (Acc. +5.9) & \textbf{0.12087} (Acc. +6.8) & \textbf{0.06772} (Acc. +1.13) \\
    % Acc. {$\uparrow$} & 5.9 & 6.8 & 1.13 \\
    \midrule
    \debertabase{} & 0.19735 & 0.13267 & \textbf{0.05191} \\
    \debertacl{} & \textbf{0.20029} (Acc. +0.6) & \textbf{0.14343} (Acc. +3.7) & 0.05126 (Acc. +0.53) \\
    % Acc. {$\uparrow$} & 0.6 & 3.7 & 0.53 \\
    \bottomrule
    \end{tabular}
    \caption{Variance in class-level accuracy (accuracy increase by each contrastive model is listed for reference). The higher the variance, the more the model performance varies between different classes. For each dataset, higher variance for each baseline/contrastive pair is \textbf{bolded}.}
    % \vspace*{4in}
    \label{table:class_variance}
\end{table*}

\newpage

\newpage

\end{document}